\documentclass[10pt,conference]{IEEEtran}

\usepackage[colorinlistoftodos]{todonotes}
\usepackage[english]{babel}
\usepackage[utf8]{inputenc}
\usepackage{amsmath} 
\usepackage{authblk} 
\usepackage{booktabs}
\usepackage{float} 
\usepackage{fullpage} 
\usepackage{graphicx}
\usepackage{graphicx} 
\usepackage{subcaption} 
\usepackage{setspace} 
\usepackage{url}
\usepackage{verbatim}
\usepackage{multirow}
\usepackage{rotating}

\title{A New Urban Objects Detection Framework Using Weakly Annotated Sets}

\author[1]{Eric Tokuda}
\author[1]{Gabriel A. Ferreira}
\author[2]{Claudio Silva}
\author[1]{Roberto M. Cesar-Jr}

\affil[1]{University of São Paulo - USP\\ São Paulo - Brazil }
\affil[ ]{\textit{\{keiji, gabriel.augusto.ferreira, rmcesar\}@usp.br}}

\affil[2]{New York University}
\affil[ ]{\textit{csilva@nyu.edu}}

\begin{document}
\maketitle

\begin{abstract}
Urban informatics explore data science methods to address different urban issues intensively based on data. The large variety and quantity of data available should be explored but this brings important challenges. For instance, although there are powerful computer vision methods that may be explored, they may require large annotated datasets. In this work we propose a novel approach to automatically creating an object recognition system with minimal manual annotation. The basic idea behind the method is to use large input datasets using available online cameras on large cities. A off-the-shelf weak classifier is used to detect an initial set of urban elements of interest (e.g. cars, pedestrians, bikes, etc.). Such initial dataset undergoes a quality control procedure and it is subsequently used to fine tune a strong classifier. Quality control and comparative performance assessment are used as part of the pipeline.  We evaluate the method for detecting cars based on monitoring cameras. Experimental results using real data show that despite losing generality, the final detector provides better detection rates tailored to the selected cameras. The programmed robot gathered 770 video hours from 24 online city cameras (\~300GB), which has been fed to the proposed system. Our approach has shown that the method nearly doubled the recall (93\%) with respect to state-of-the-art methods using off-the-shelf algorithms.  

\end{abstract}

\section{Introduction} 


Modern cities collect a vast amount of data daily~\cite{united2016air}, which includes  information about mobility, energy, violence, pollution and cultural life, to name but a few. There is a huge amount of data available and processing it is not an easy task. Particularly, much information can be deduced when multiple sources of information are considered. Image and video sources are particularly very rich ones. Useful information is not always promptly available from the data. In some cases, great manual effort is necessary to process and combine different data sources to obtain the desired information. 

\begin{figure}[ht!]
    \centering
    \includegraphics[width=0.45\textwidth]{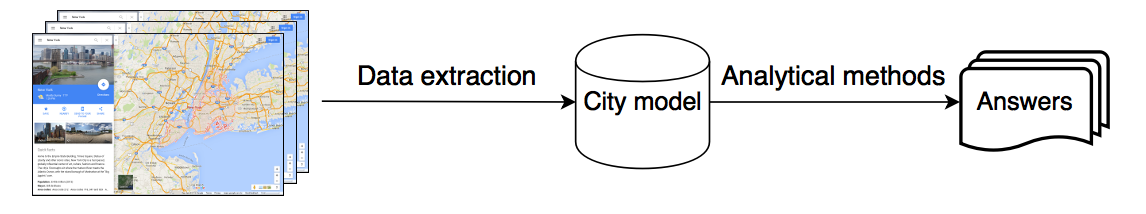}
    \caption{Urban informatics framework under development. Data is acquired from different sources, processed and integrated to compose a city a model and address relevant urban issues.}
    \label{fig:workflow}
\end{figure}

\begin{figure}[ht!]
    \centering
    \begin{subfigure}[b]{0.4\textwidth}
        \includegraphics[width=\textwidth]{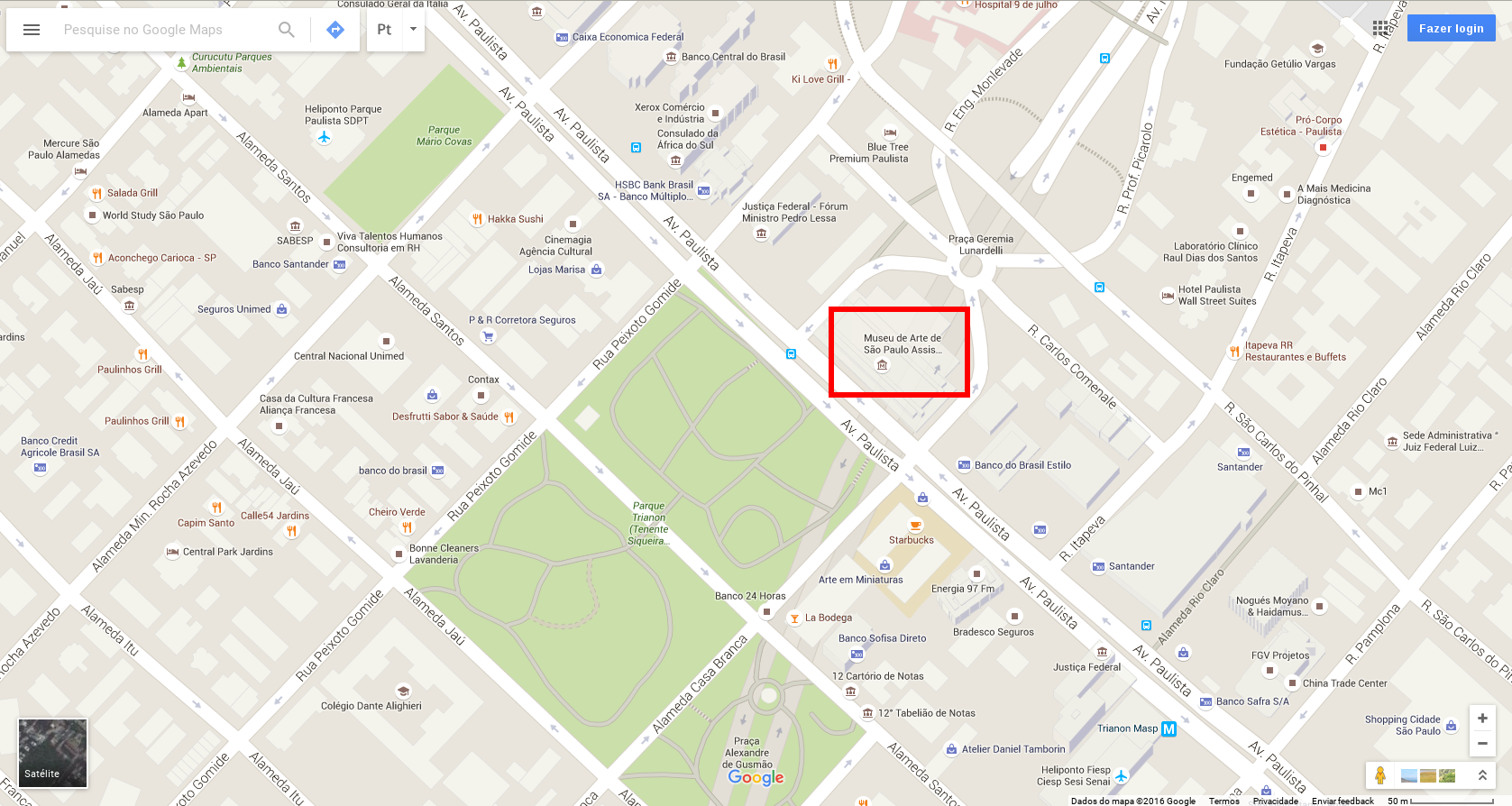}
        \caption{City map}
    \end{subfigure}
    \begin{subfigure}[b]{0.4\textwidth}
        \includegraphics[width=\textwidth]{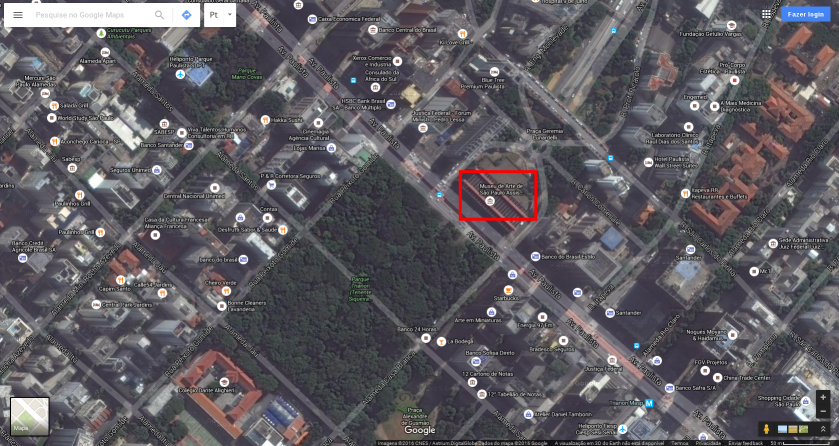}
        \caption{Remote sensing images}
    \end{subfigure}
    \begin{subfigure}[b]{0.4\textwidth}
        \includegraphics[width=\textwidth]{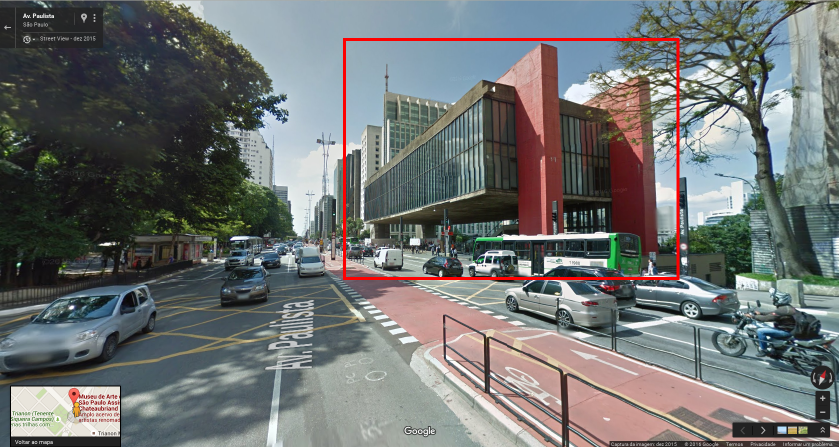}
        \caption{City images}
    \end{subfigure}

    \caption{Three different visual data sources: images from cartography, from remote sensing and from monitoring cameras. The combination of them constitutes an important feature in the creation of a city model. The red selection exhibits the same urban element linked throughout the three different classes of images.}
        \label{fig:sources}
\end{figure}

In this paper, we describe an ongoing project based on a framework to automatically combine different sources of information and to create a city model to address urban issues (Figure~\ref{fig:workflow}). One broad categorization of the data is in visual and non-visual data. Visual data, in turn, accommodates at least three categories that are relevant to our work that are: city maps, remote sensing images and street-level images. Particularly, city images contain relevant information regarding urban elements including cars, people and buildings, and it is useful to have all of them identified in the data. With the combination of them, one would be able to obtain geographic coordinates based on the recognition of a building which position is known~(Figure~\ref{fig:sources}). In this stage of our framework, we focus on the recognition of cars in city images, but an extension to other urban elements is direct. Extracting semantic information of visual data is challenging and may require great manual effort. An option is to use crowd-sourcing services like \emph{Amazon Mechanical Turk}~\cite{mturk2013}. It works reasonably well but the quality may fall short depending on the kind of task~\cite{sorokin2008utility}.  Our contribution is the proposal of a new method of automatic generation of objects detectors, with no manual annotation. We evaluate the approach in the task of creating a car detector for monitoring cameras images.

We propose an approach to automatically generate object detectors for urban informatics. In our method, data is harvested, pre-annotated with a weak classifier and then used to fine-tune a model. We validated it through the creation of a car detector for video sequences captured in the wild from monitoring cameras.

This paper is organized as follows. The remaining of this section presents relevant works to this approach. Section~\ref{sec:proposed} explains in details the method proposed while Section~\ref{sec:experimental} presents the experimental results and validation of the method. The paper is concluded in Section~\ref{sec:conclusion} with final remarks on ongoing work.

\subsection{Related Work}
\label{subsec:related}

Object recognition may involve the detection of objects in a scene~\cite{szeliski2010computer}. Due to the emergence of rich datasets and the development of higher computing capabilities, object recognition has been one prominent task in computer vision. Meaningful advances have been achieved since the theme started to be explored (\cite{viola2001rapid,lowe1999object,felzenszwalb2010object}) and methods such as~\cite{ren2015faster} allow fast and high accuracy recognition rates. When we restrict the recognition to one known class, then we have a problem of object detection~\cite{szeliski2010computer}.

The Deformable Parts Model (DPM) is a notable object detection method proposed by~\cite{felzenszwalb2010object}. It identifies an object through its constituent parts and the corresponding spatial dispositions. The model characterizes each part of the object through a Histogram of Oriented Gradients (HOG)~\cite{dalal2005histograms} in pyramid levels and through the possibly deformed positions relative to the root object. The overlapping candidates are computed and ranked according to a score.

Among other approaches, Artificial Neural Networks (ANNs)~\cite{schmidhuber2015deep} have being widely used for object recognition. An ANN can be thought as a parallel processor composed of simple processing units, the neurons. Each neuron processes the inputs through an activation function and sends the results to the following neurons~\cite{schmidhuber2015deep}. Neurons are organized in layers and architectures with more than one hidden layer are referred as \emph{deep}~\cite{schmidhuber2015deep} networks. In a conventional multiple layers ANN, each layer is fully connected to the preceding layer. A variant, known as Convolutional Neural Networks (CNN), comprises one or more partially connected layers. This type of ANN has many applications in image classification~\cite{krizhevsky2012imagenet,simonyan2014very,schmidhuber2015deep}. In object recognition, one direct approach is to scan every possible window of the target image and classify it independently. This solution is compute-intensive and instead of performing exhaustive search of the objects, many mechanisms of a previous step, proposals of regions of interest have been created~\cite{hosang2015makes}. They are commonly categorized in superpixels grouping (\cite{van2011segmentation,carreira2010constrained}) and variants of sliding-windows (\cite{cheng2014bing,zitnick2014edge}). Region-proposal methods are analogous to interest point detectors~\cite{hosang2015makes} because they allow focusing attention on specific regions for subsequent tasks.

The work of Regions with Convolutional Neural Networks (RCNN)~\cite{ren2015faster} introduces a unified network that performs  region proposal and  classification, which means that during the training step it accepts annotations of multiple sized objects and, during the testing stage, it performs classification of those objects in images of arbitrary sizes.


The conventional creation of an image recognition method involves  training and evaluation of the method proposed in a particular set. There are several datasets available, some covering multiple classes~\cite{griffin2007caltech,everingham2010pascal,russakovsky2015imagenet,lin2014microsoft}  while others focus in one particular class~\cite{ali2007lagrangian,dollar2012pedestrian,soomro2012ucf101}. In particular there are several car datasets available~\cite{carbonetto2008learning, yang2015large, lichman2013uci, ozuysal2009pose, krause20133d} which include great number and diversity of samples. However, they lack diverse situations such as high occlusion, varying quality, naturally moving objects and diversified image acquisition conditions such as weather, common situations in monitoring cameras.  Monitoring cameras datasets, such as Virat~\cite{oh2011large}, Kitti~\cite{geiger2013vision}, Visor~\cite{vezzani2008visor}, i-Lids~\cite{hosmer2007ilids}, CamVid~\cite{brostow2009semantic} and MIT Traffic Dataset~\cite{wang2007unsupervised} provide just limited amounts of this type of videos. Another possibility is to acquire data from public streaming. There are multiple platforms that aggregate monitoring cameras from many places around the world, like EarthCam~\cite{earthcam}, InseCam~\cite{insecam} and Camerite~\cite{camerite2016}. Due to their nature, such image sources present scenes with varying conditions, including scenes with scarce illumination, low resolution and bad weather conditions.

In the context of automatic learning of concepts, the work of~\cite{misra2015watch} introduces a framework addressing the semi-supervised learning problem for discovering multiple objects in sparsely labeled videos. Focus is given to the automatic quality assurance, due to the quick worsening of the classifier when false positives are included in the database. The method starts with few sparsely labeled videos and exemplar detectors~\cite{hariharan2012discriminative} are trained on these starting data. The video is consistently sampled and annotated by the classifier. Since the annotations are sparse, the authors argue that  negative examples cannot be obtained from the neighborhood of a detection, so the use random image from external sources. An initial filtering is performed using temporally consistency, assuming a smoothness in the motion of the objects. Then an outliers removal approach was applied in a different feature space from the classifier using the unconstrained Least Squares Importance Fitting. The filtered detection serve as starting point for a short-term tracking, using sparse optical flow using HOG features. To filter the potentially redundancy in the resulting  set, each object is associated with the exemplar detector~\cite{hariharan2012discriminative} and high correspondences, corresponding to redundant detections, are removed. Finally, the final data is used to update the detectors. The current work here presented has many similarities to this work. Next we explain in detail the steps of our approach.


\section{Proposed Framework}
\label{sec:proposed}

\begin{figure*}
    \centering
    \includegraphics[width=0.8\textwidth]{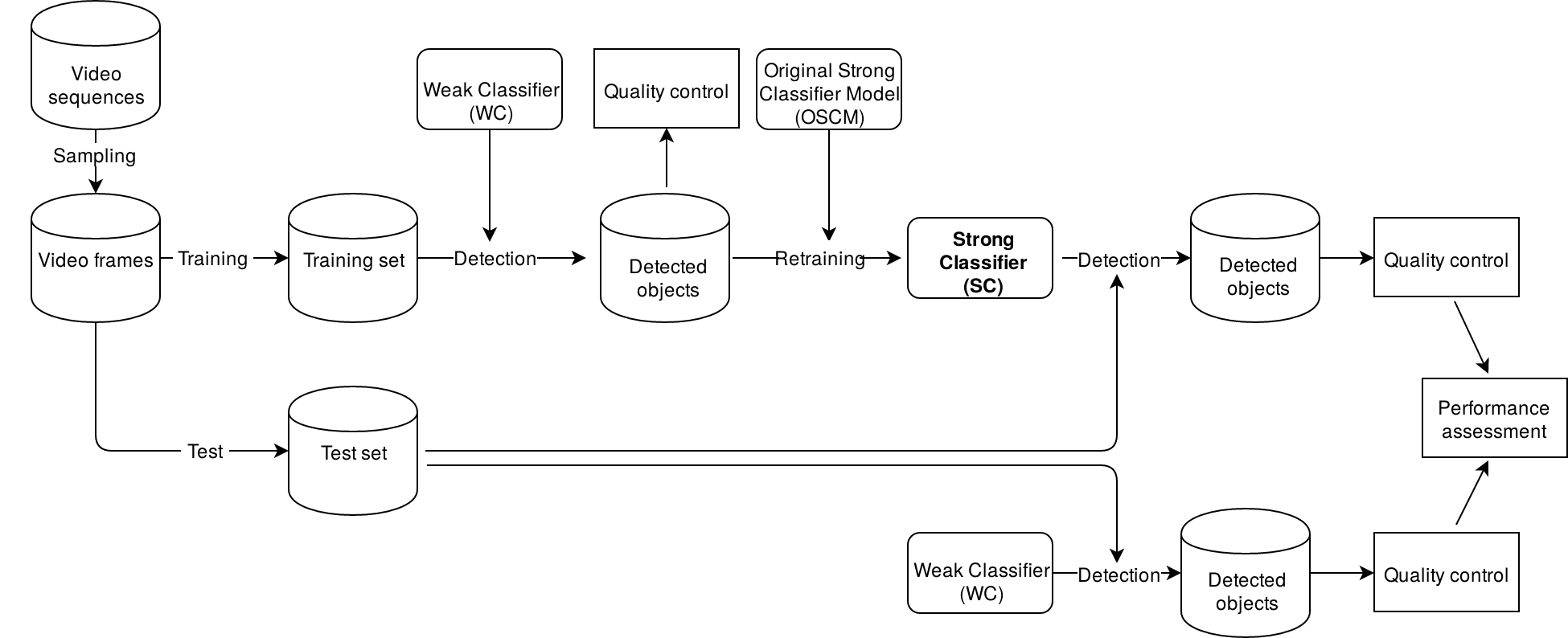}
    \caption{Proposed methodology. Frames are sampled and split in training and test sets. A weak detection is performed on the training set and used to retrain the strong classifier. Quality control is performed both on the detections of the new dataset on test set and also on the detections of the weak classifier on the same set.}
    \label{fig:method}
\end{figure*}

Figure~\ref{fig:method} describes the introduced method. We propose to create a detector with minimal user intervention, provided a dataset and two different classifiers. They are referred as Weak Classifier ({\bf WC}) and Original Strong Classifier Model ({\bf OSCM}). The resulting fine-tunned classifier is referred as  Strong Classifier ({\bf SC}).The method steps shown in Figure~\ref{fig:method} are described below. 


\subsection{Data acquisition}

\begin{figure*}
\centering
\begin{tabular}{cc}
 \includegraphics[width=0.45\textwidth]{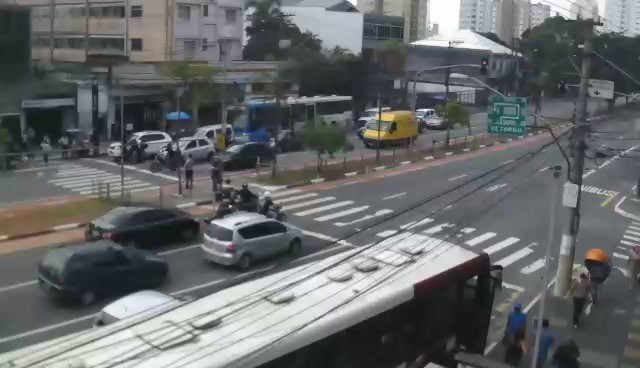} &   \includegraphics[width=0.45\textwidth]{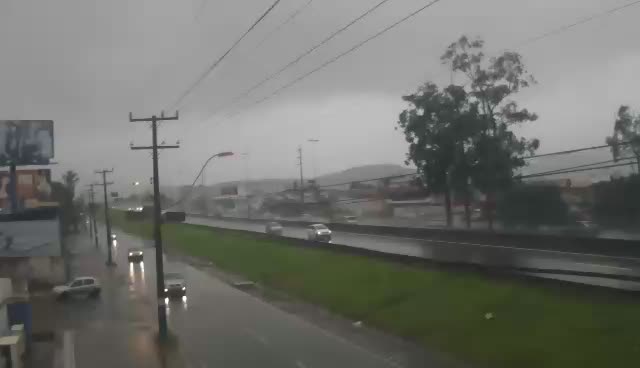} \\[6pt]
 \includegraphics[width=0.45\textwidth]{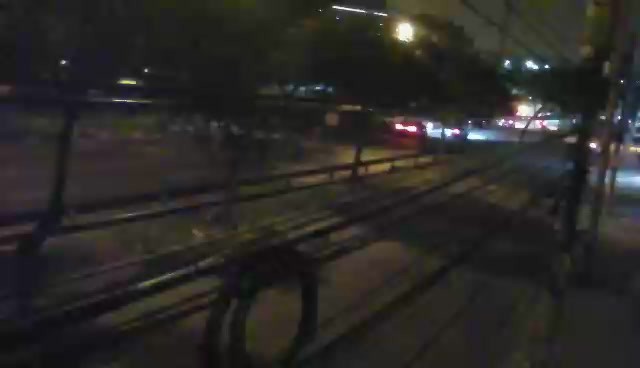} &   \includegraphics[width=0.45\textwidth]{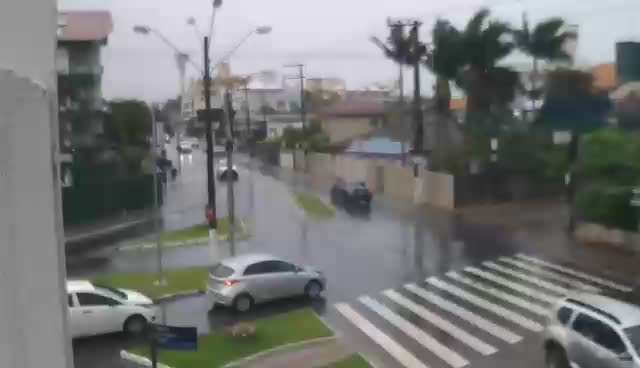} \\
\end{tabular}
   \caption{Data source. Sample of the video frames used.}
   \label{fig:sample}
\end{figure*}

The first stage  consists in the acquisition of a large set of images and video of interest. Besides using available datasets, the proposed implementation retrieves data from monitoring cameras available. This solution allows an excellent variability and amount of data that may be retrieved. We used~\cite{camerite2016} to filter and select the monitoring cameras. See Figure~\ref{fig:sample} for some sample frames obtained.

\subsection{Data sampling}
An optional step in the method is sampling the input video. The motivation is the redundancy of the objects that might be detected in consecutive frames, if no temporal coherency such as ~\cite{kalal2012tracking} is taken into account. This step is also important to cope with the large datasets possibly obtained in the previous step.

\subsection{Training and test sets division}
To properly evaluate the method proposed, the data should be split into training and test set. A $10\%$ to $20\%$ are rules-of-thumb commonly accepted in the field~\cite{abu2012learning}. This allows performance assessment using cross-validation like strategies.

\subsection{Weak classifier application and fine tunning training set generation}

The DPM method~\cite{felzenszwalb2010object} was adopted as our WC. DPM is a remarkable work in object recognition with the use of low level features. The model assumes that an object is composed of components, each of which composed by a root and a set of parts. Each object hypothesis is computed as Equation~\ref{eq:dpmscore}, the difference of a filter and deformation term plus a bias. The score of the filter part is given by the convolution of the model $F_i$ convolved with the HOG features extracted at their own location. The deformation term consists in the convolution of the model deformation parameters and the displacement of the part. An object hypothesis is considered a detection if it results in the best placement of parts, as expressed by Equation~\ref{eq:dpmmatching}. The training set is processed by the WC and a set of objects detections is obtained (Figure~\ref{fig:method}).  

\begin{align}
\label{eq:dpmscore}
&score(p_0,\ldots, p_n) = \nonumber \\ &=\sum_{i=0}^n F_i^{\prime } \cdot \phi (H, p_i) - \sum_{i=1}^n d_i \cdot \phi_d(dx_i, dy_i) + b,
\end{align}

\begin{align}
\label{eq:dpmmatching}
score_(p_0) = \max\limits_{p_0, \ldots, p_n} score(p_0, \ldots, p_n) ,
\end{align}


\subsection{Strong classifier generation}
The detected objects are used to fine-tune a RCNN pre-trained on ImageNet. RCNN in its original form~\cite{girshick2014rich} performs a ordinary classification in each region of the image. The feature extraction is performed using a CNN and the classification is performed using Support Vector Machines~\cite{cortes1995support}. An additional step of region proposals can be added to cut out the search space of objects detection in the image. A relevant drawback of this original version is that it is very compute intensive. A faster version of the RCNN has been proposed~\cite{ren2015faster}. It starts by applying the convolutional layers followed by region proposals extraction. Classification is then performed as the last step. The resulting method is faster and it presents similar results~\cite{ren2015faster}. The fine-tunned RCNN represents a final detector Strong Classifier tailored for the input online city cameras.

\subsection{Quality control}

A critical aspect of the proposed approach is to assure the quality of the intermediate representations obtained by semi-automated ways. In the approach described in last section, the weak classifier is used to generate samples that are fed to train and fine-tune the strong classifier.  A quality control step is performed to evaluate the performance of WC in this task.

\begin{figure}[ht!]
    \centering
    \includegraphics[width=0.4\textwidth]{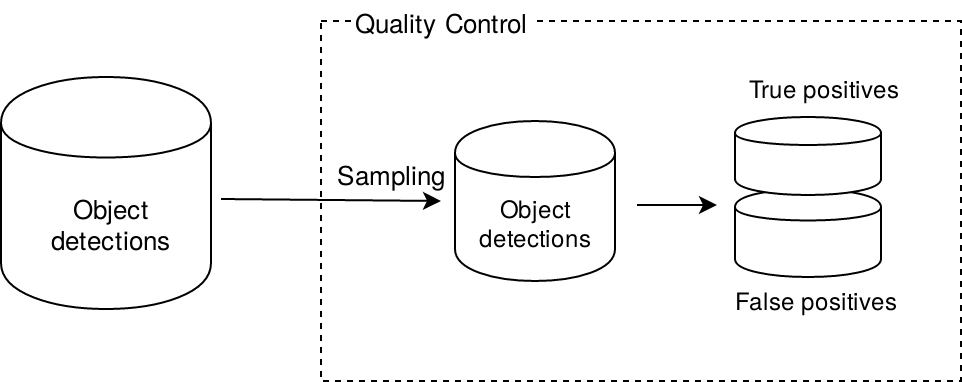}
    \caption{Quality control methodology. In the quality control stage, a manual inspection is performed on the detections and the number of true positives and false positives are computed.}
    \label{fig:qualitycontrol}
\end{figure}

In the quality control stage (Figure~\ref{fig:qualitycontrol}) we want to estimate the real performance of an object detector given the detected objects in a sample.

In order to evaluate the accuracy of each detector over the  dataset, it is desirable to have the proportion $p$ of the true-positives each detector has produced. However, to avoid excessive manual work on annotating the images, we estimate the proportion $\hat{p} = TP/(TP+FP)$ over a randomly selected sample from the dataset.  The confidence interval for a population proportion $\hat{p}$ based on a sample of size $n$ is given by~\cite{statisticsForComputer}:

$\hat{p} \pm z_{\alpha/2} \cdot \sqrt[]{\frac{\hat{p}(1-\hat{p})}{n}} $,

\noindent where $z$ is the normal distribution.

We have to use a reasonable $\hat{p}(1-\hat{p})$ value, based on our experience and on which outcome we expect from this experiment. The worst case (when we need the largest $n$) happens if $\hat{p} = 0.5$, since $\hat{p}(1-\hat{p})$ reaches its maximum value. A good practice is to collect a small random pilot sample and to calculate the proportion over this sample. Based on our experiences and on a pilot sample of 50 images, we chose to use $\hat{p}(1-\hat{p})=(0.8 \times 0.2)$. Thus, with 95\% confidence ($z_{0,025} = 1.96$), and a margin of error $\epsilon \approx 2.7\%$, we need $n \approx 850$.

\begin{equation}
	\label{eq:prec}
	precision = \frac{TP}{TP + FP}
\end{equation}

\begin{equation}
	\label{eq:recall}
	recall = \frac{TP}{TP + FN}
\end{equation}

The precision on the sample is computed according to Equation~\ref{eq:prec}. The recall of the population is in principle unknown. To be calculated, it requires the number of false negatives (see Equation~\ref{eq:recall}). Once again, instead of annotating all the dataset for recall, the same approach for estimating the precision can be done, by annotating a randomly selected sample.

The performance evaluation of the SC is carried out based on the results of the two quality control stages. Despite we want to avoid computing the recall of each classifier (since it requires manual annotation of a very large video set), the number of images tested for both WC and SC are the same, i.e., $TP_{WC}+ FN_{WC} = TP_{SC}+ FN_{SC}$, so we can compute the relative change in the recall of WC, $rc(recall_{WC})$ as Equation~\ref{eq:increase}. This is particularly interesting in the big data scenario, where we want to minimize the effort to label the data.

\begin{align}
	\label{eq:increase}
	&rc(recall_{WC}) =\nonumber\\
    &= \frac{recall_{SC} -recall_{WC} }{recall_{WC}} \nonumber\\
    &= \frac{ \frac{TP_{SC}}{TP_{SC}+ FN_{SC}} - \frac{TP_{WC}}{TP_{WC}+ FN_{WC}}} {\frac{TP_{WC}}{TP_{WC}+ FN_{WC}}} \nonumber\\
    &= \frac{TP_{SC}- TP_{WC}}{TP_{WC}}
\end{align}


The quality control stage can be summarized in the following sequence of steps:
\begin{enumerate}
\item Compute a sample size, according to the equation of minimum sample size.
\item Label a small sample. 
\item Compute TP, TN and FP.
\item Apply Equation~\ref{eq:increase}.
\item Compute FN and Equation~\ref{eq:recall}.
\end{enumerate}

The last step is optional in case one just needs the relative improvement of the method, which can be obtained through Equation~\ref{eq:increase}. In case this step is performed, the false negative samples can be used in the retraining of the SC, as proposed by~\cite{felzenszwalb2010object}.

\subsection{Comparative performance assessment}
The SC detector is evaluated on the test set and a quality control stage is also performed. The test set is also processed by the WC and quality control is performed in this stage as well. This allows assessing the gain obtained by the strong classifier w.r.t. the weak classifier. The results of the two quality control stages are combined to infer the accuracy of the SC.







\section{Experimental Results and Validation}
\label{sec:experimental}

The proposed methodology has been implemented in order to be validated. We used a relational database management system to store all the metadata of the acquisition and detections.  We used a PostgreSQL database~\cite{postgresql1996}. Different programming languages were used in the framework, including python, Matlab and Linux bash scripting language. In the acquisition stage, we continuously decoded HTTP live streaming~\cite{stockhammer2011dynamic} to MPEG~\cite{le1991mpeg} files. This stage included a failure-proof mechanism to take into account issues on the client-side, server-side and on the network. In the object detection stages, DPM~\footnote{https://github.com/rbgirshick/voc-dpm} and the Python implementation~\footnote{https://github.com/rbgirshick/py-faster-rcnn} of the Faster-RCNN~\cite{ren2015faster} method were used.

We validated the proposed method in two experiments. In the first case, we created a car detector for images  from the monitoring cameras. In the second case we were motivated by the task of efficiently finding cars in  rainy weather, a more difficult computer vision task. There are multiple works dealing with the problem of rain removal~\cite{zhang2006rain}, but none tackling the problem of finding cars in the rain. We restricted our dataset to images of rainy weather and developed our pipeline according to this data source.

\begin{table*}[ht]
        \caption{Comparison of the two datasets, Camerite-all and Camerite-rainy.}
        \centering
        \begin{tabular}{lcccc}
        \toprule
        \multirow{2}{*}{Dataset}	&Number of	& Total		&Number of	&\multirow{2}{*}{Size (GB)}\\ 
          							&cameras	& hours		&sampled Frames &\\ \midrule
        Camerite-all				&24			&768.0		&358,036	&262.5\\
        Camerite-rainy				&14			&63.7		&7,011		&5.8\\
        \bottomrule
        \label{tab:datasets}
        \end{tabular}
\end{table*}

\begin{figure}[h]
\centering
\begin{tabular}{c}
 \includegraphics[width=0.45\textwidth]{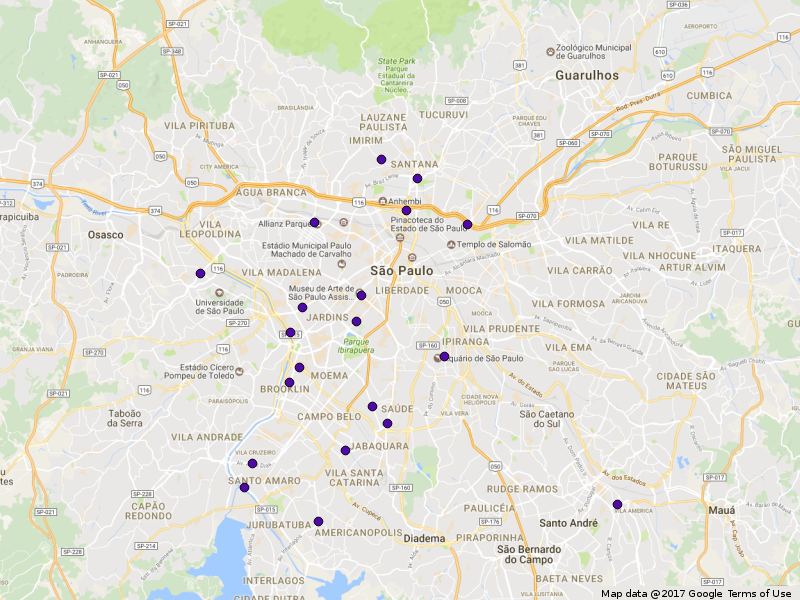}
\end{tabular}
   \caption{Locations of some of the cameras used in São Paulo city, Brazil. Map from~\cite{googlemaps}.}
\end{figure}


\begin{table*}[ht]
        \caption{Sample of the detections performed by the SC $RCNN_{all}$.}
        \centering
        \begin{tabular}{cccc}
        \includegraphics[width=0.23\textwidth]{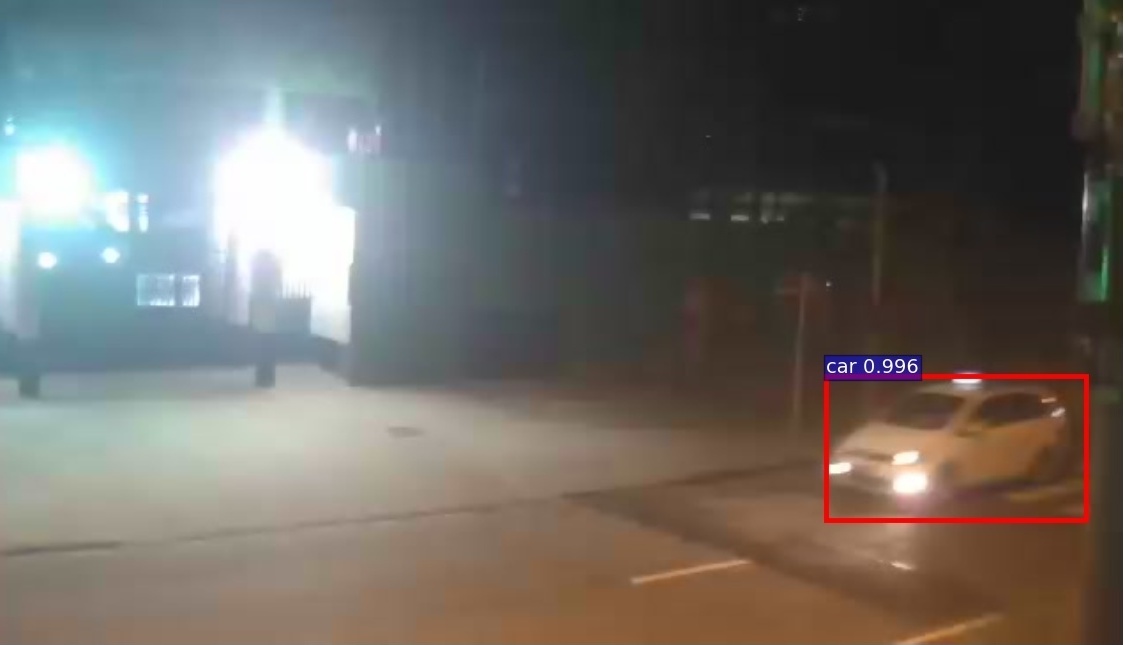} &
        \includegraphics[width=0.23\textwidth]{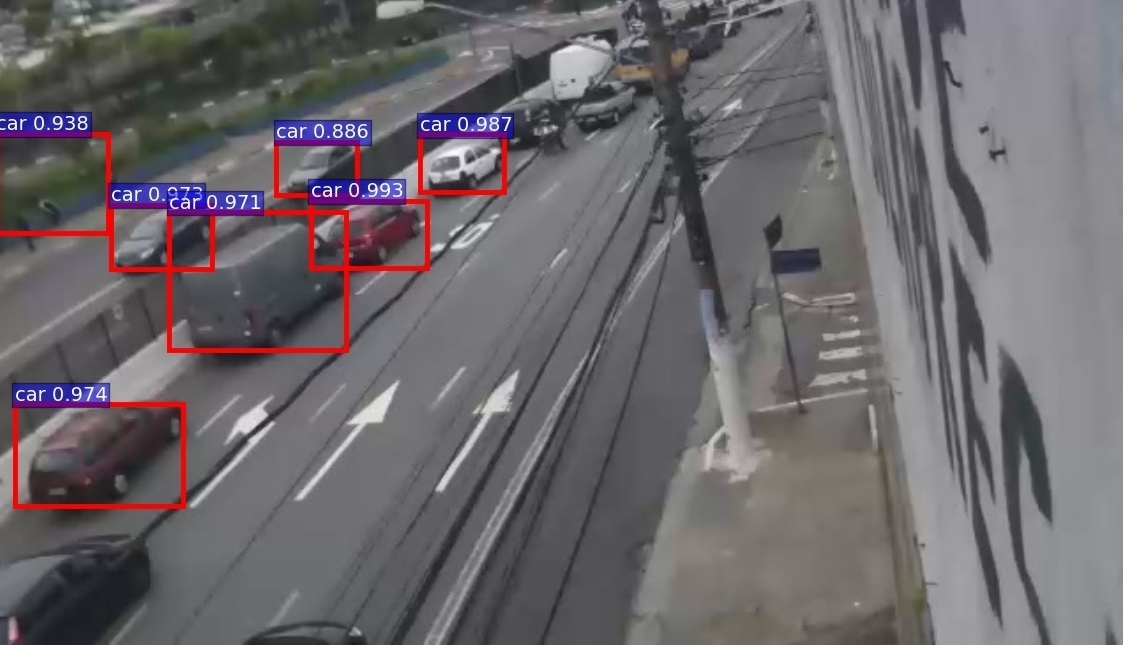} &
        \includegraphics[width=0.23\textwidth]{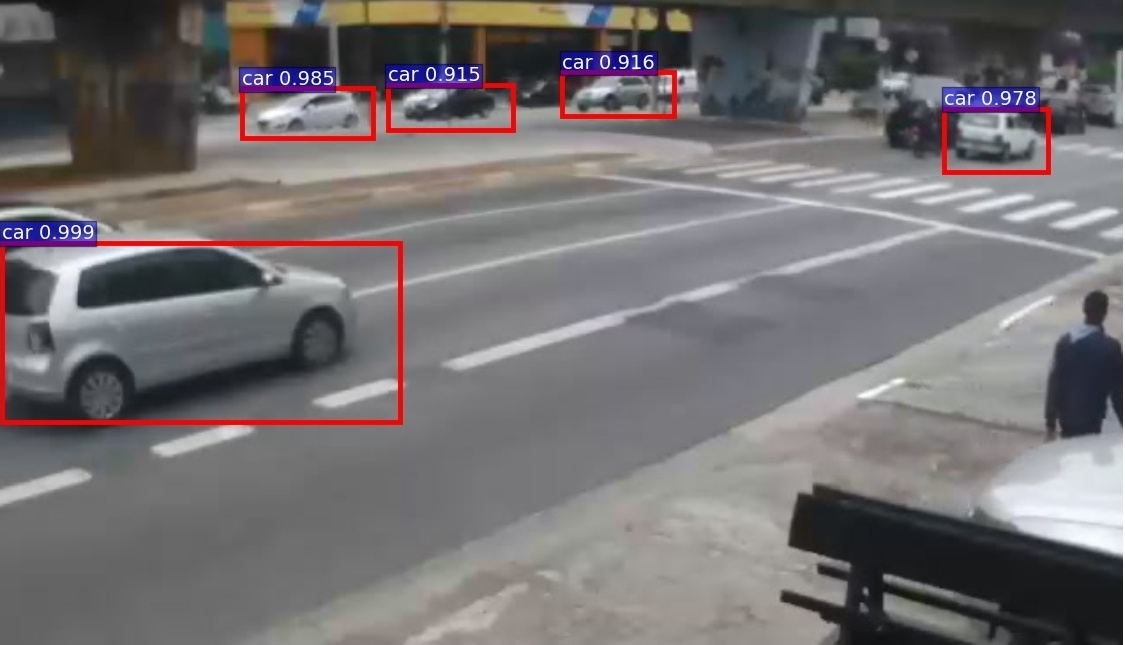} &
        \includegraphics[width=0.23\textwidth]{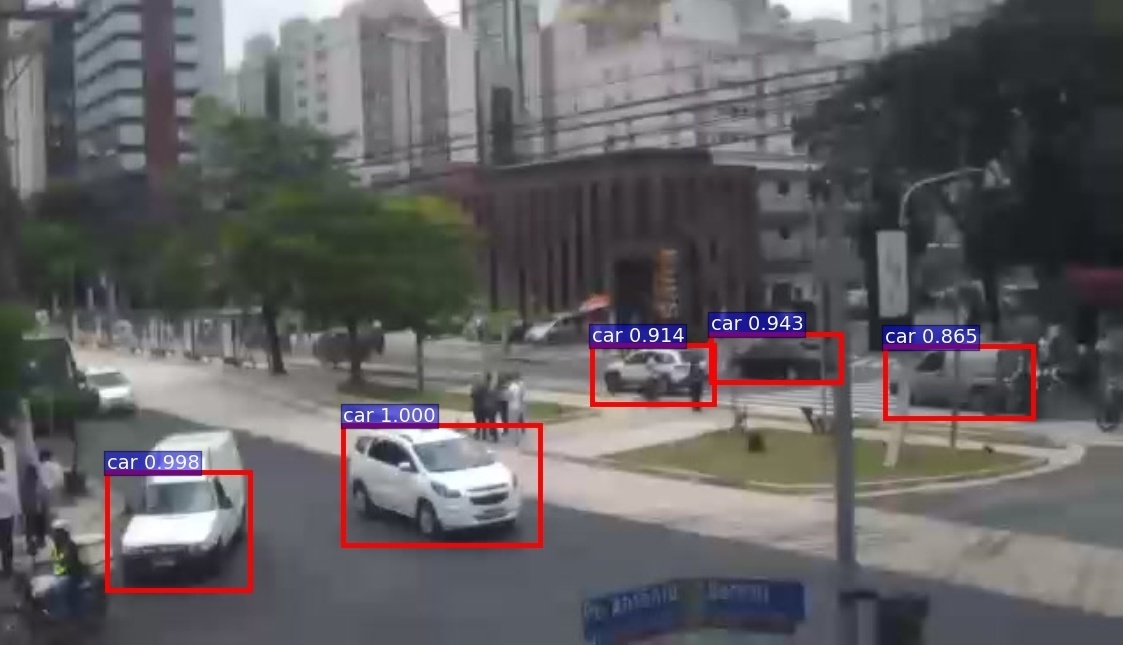} \\
        \includegraphics[width=0.23\textwidth]{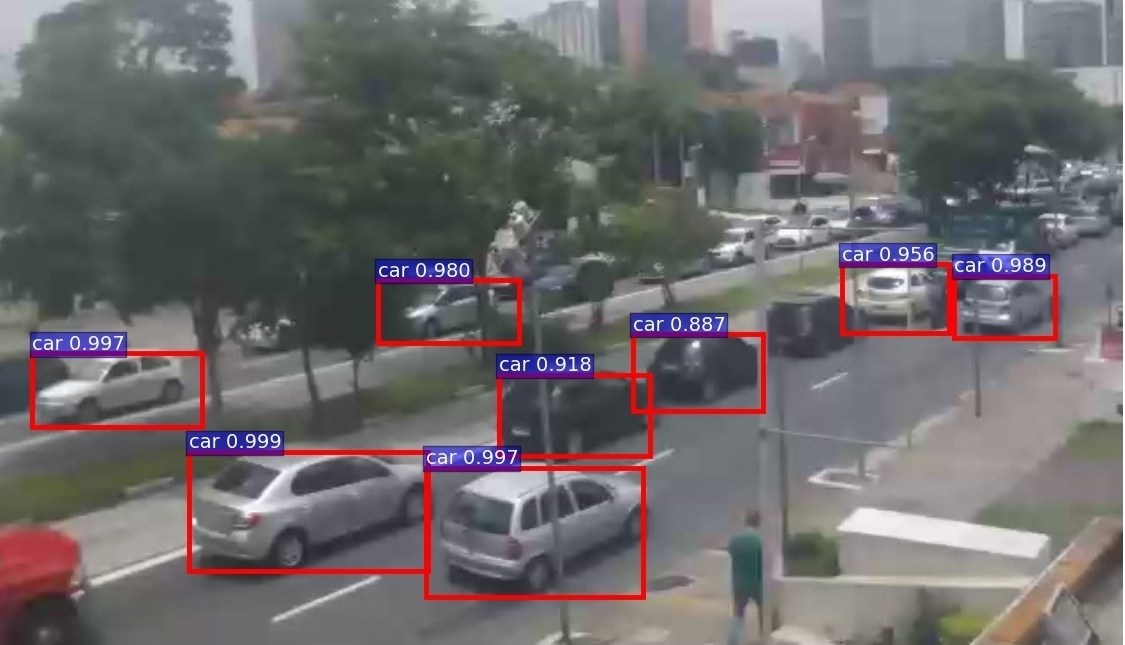} &
        \includegraphics[width=0.23\textwidth]{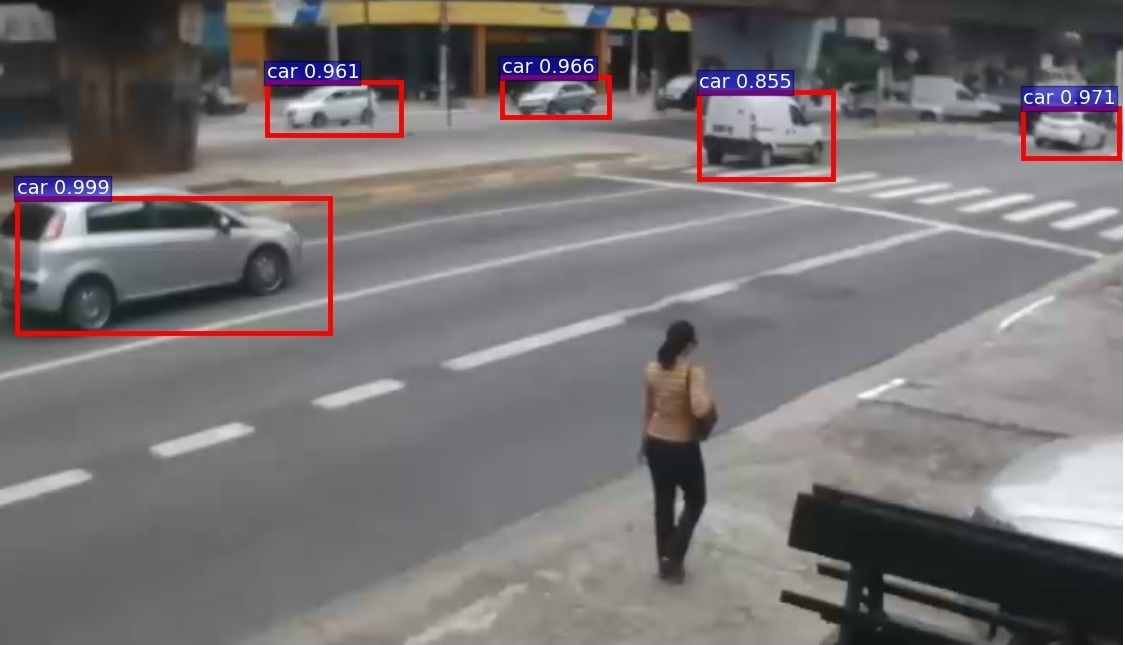} &
        \includegraphics[width=0.23\textwidth]{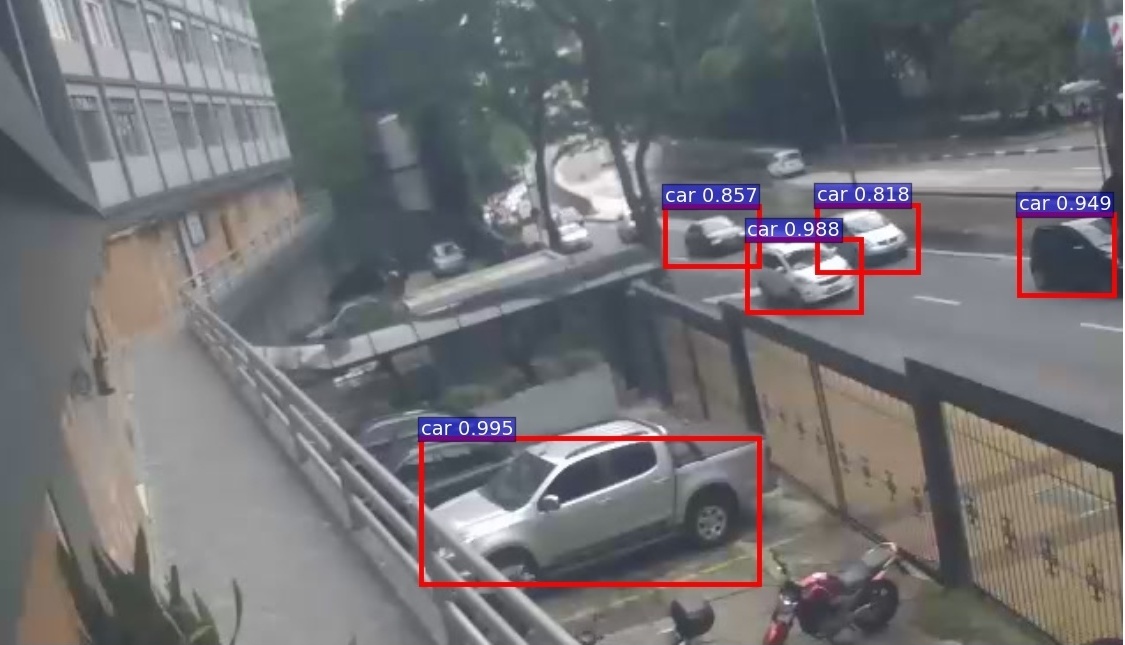} &
        \includegraphics[width=0.23\textwidth]{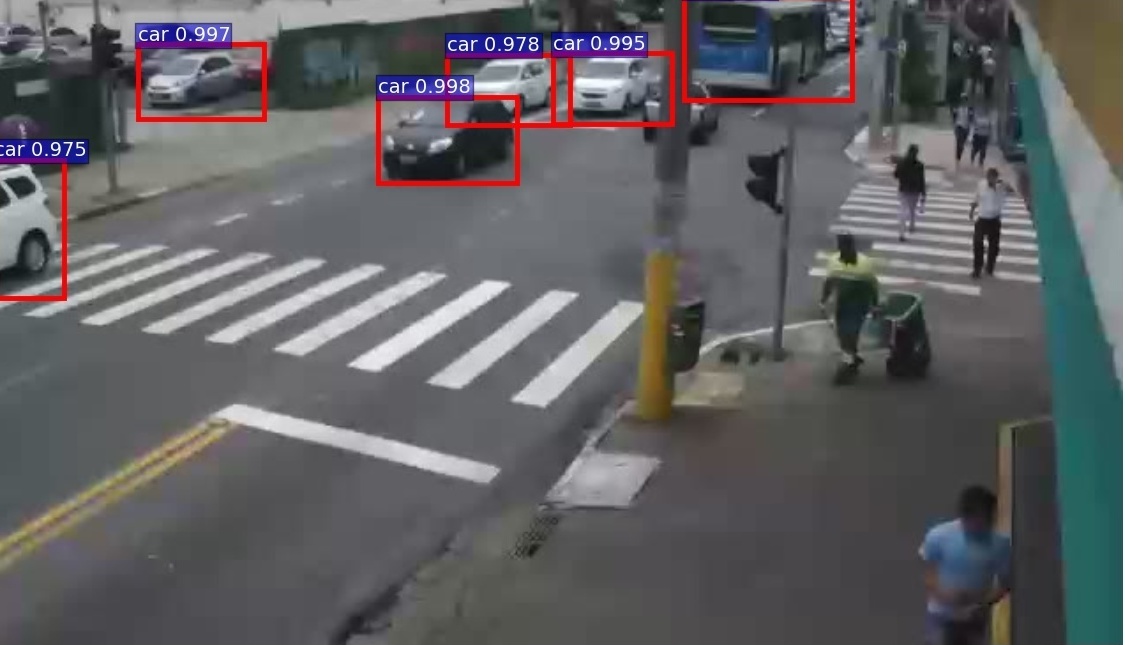} \\
        \includegraphics[width=0.23\textwidth]{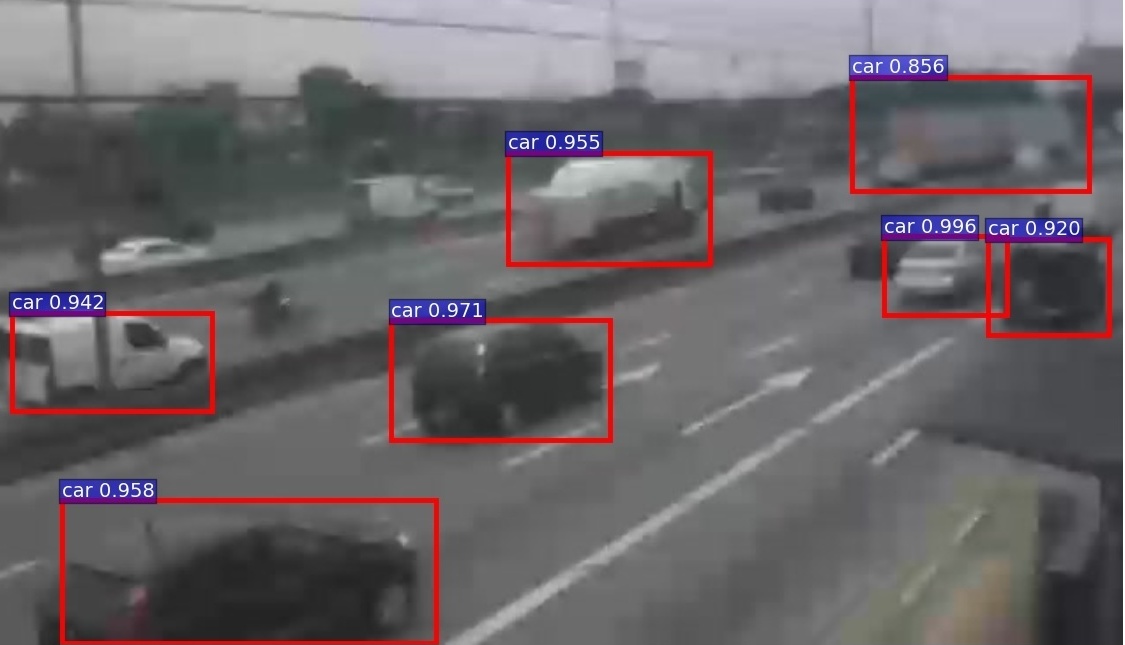} &
        \includegraphics[width=0.23\textwidth]{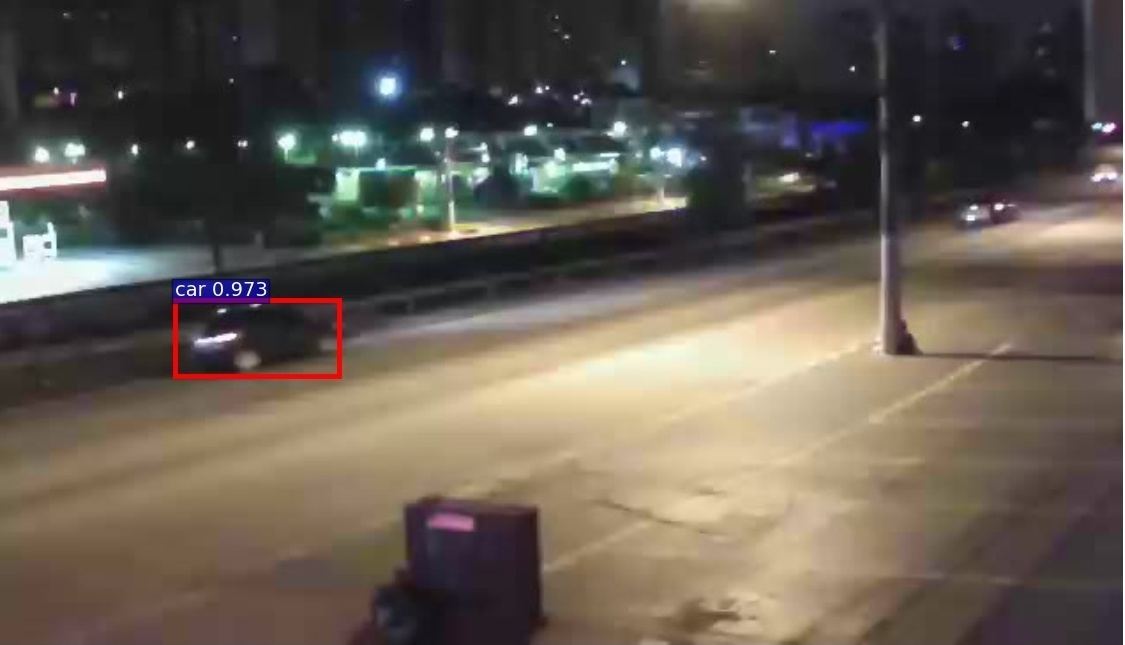} &
        \includegraphics[width=0.23\textwidth]{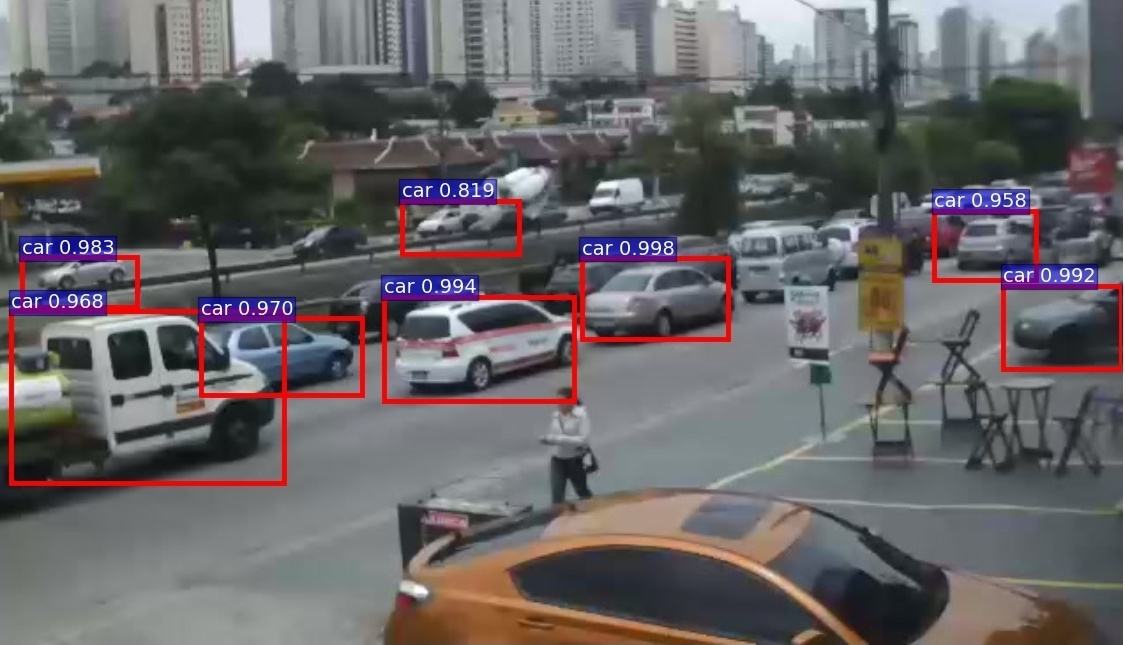} &
        \includegraphics[width=0.23\textwidth]{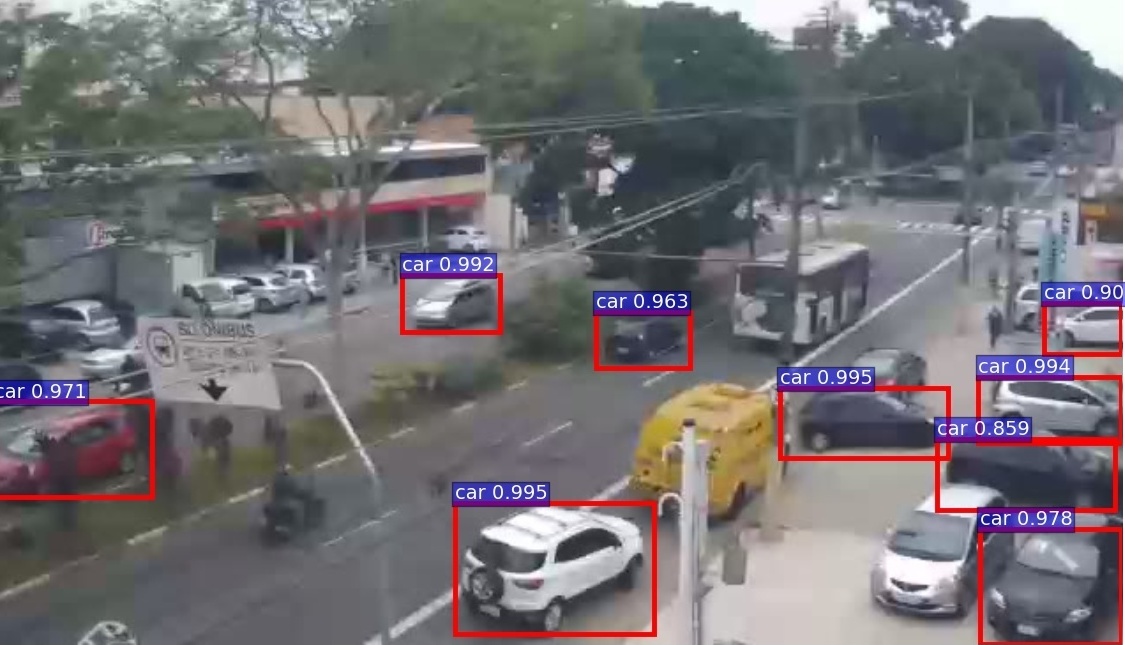} \\
        \includegraphics[width=0.23\textwidth]{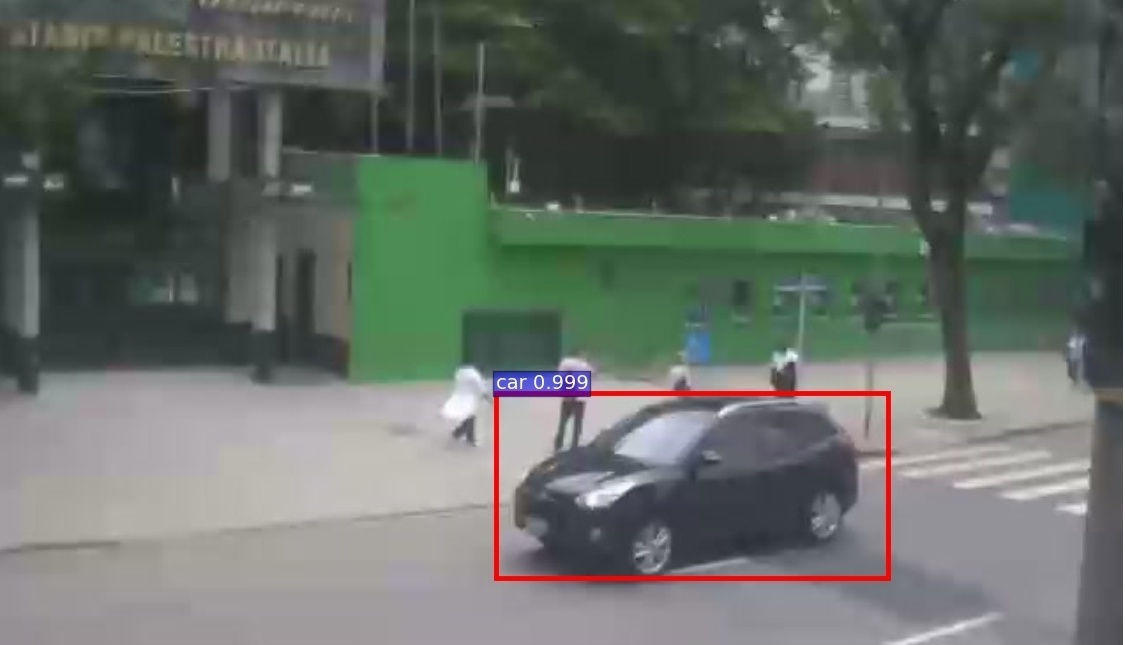} &
        \includegraphics[width=0.23\textwidth]{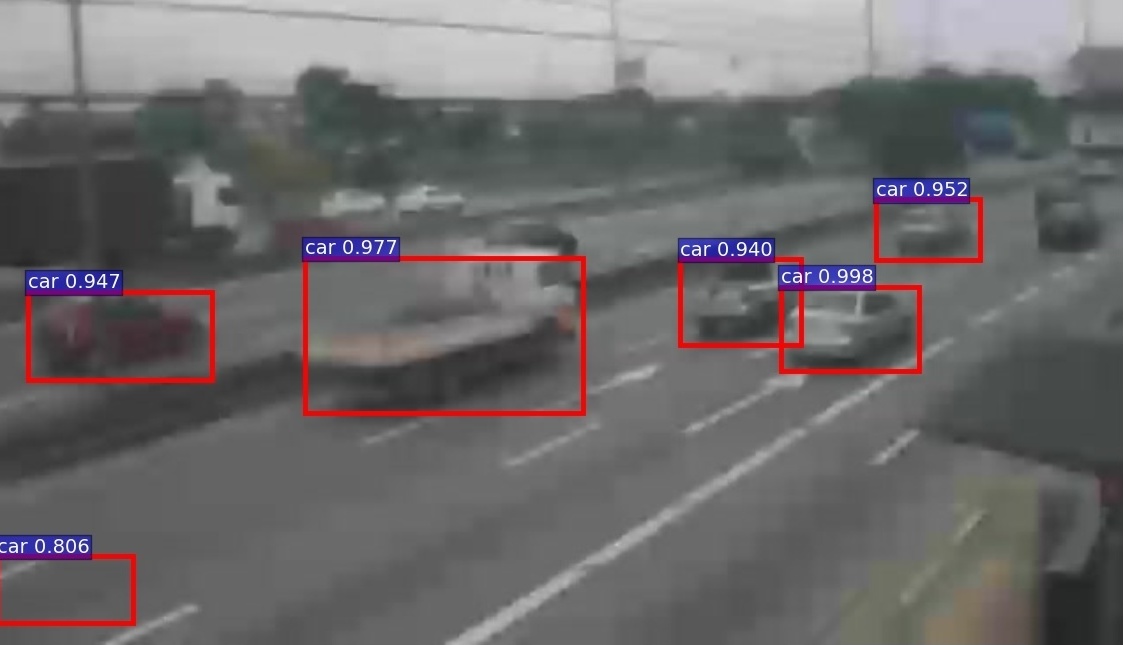} &
        \includegraphics[width=0.23\textwidth]{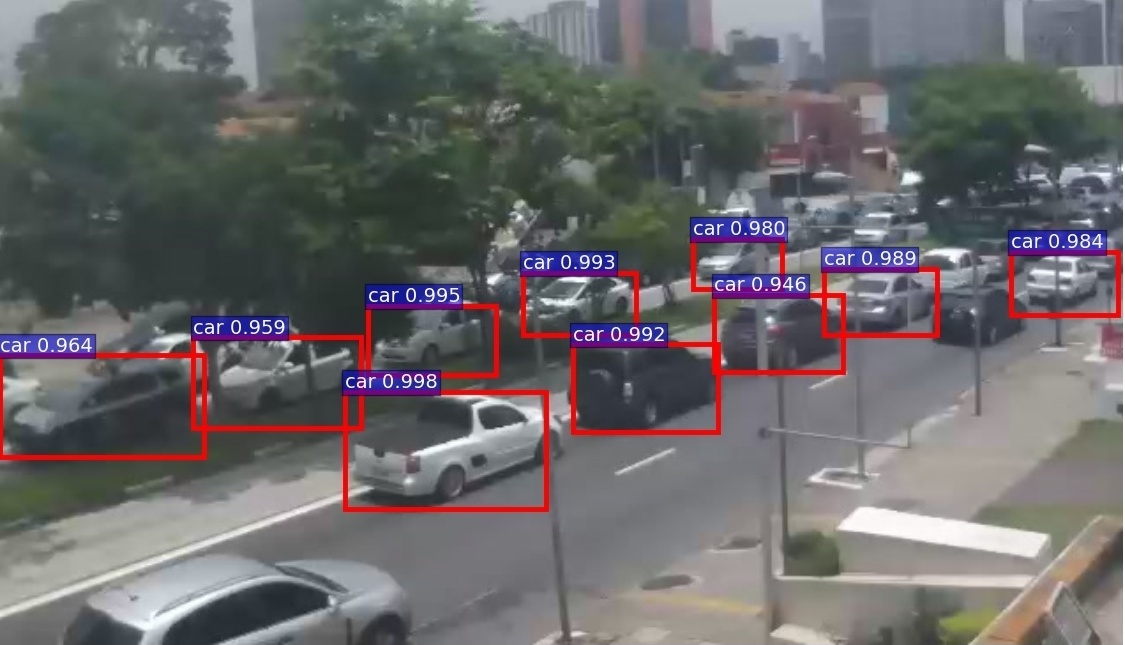} &
        \includegraphics[width=0.23\textwidth]{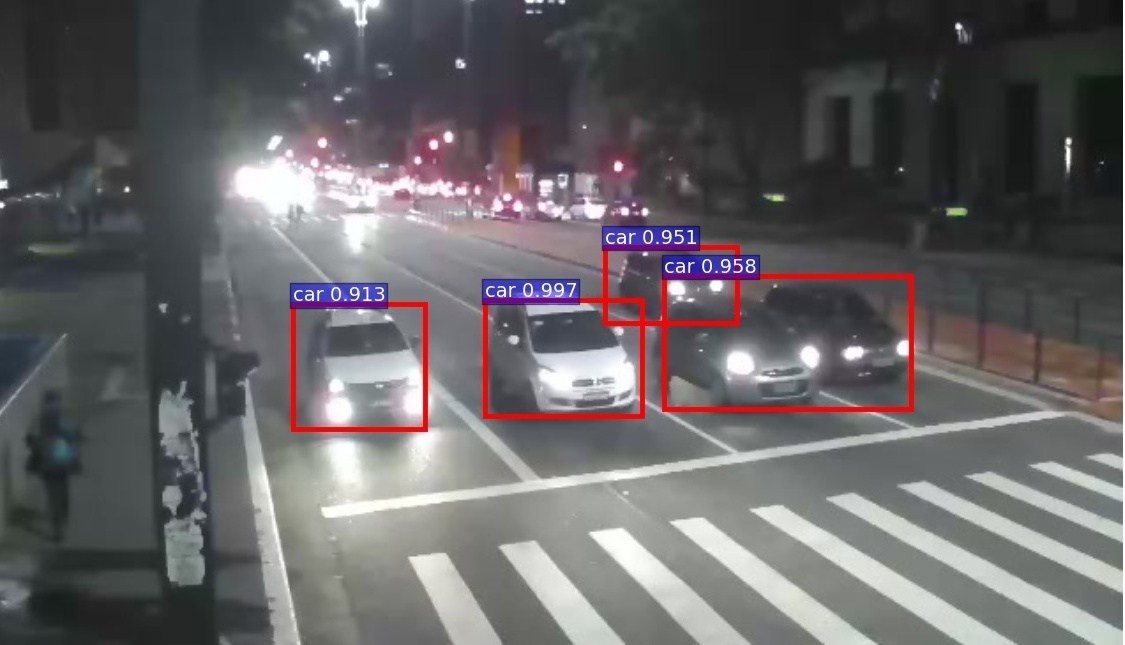} \\
        \includegraphics[width=0.23\textwidth]{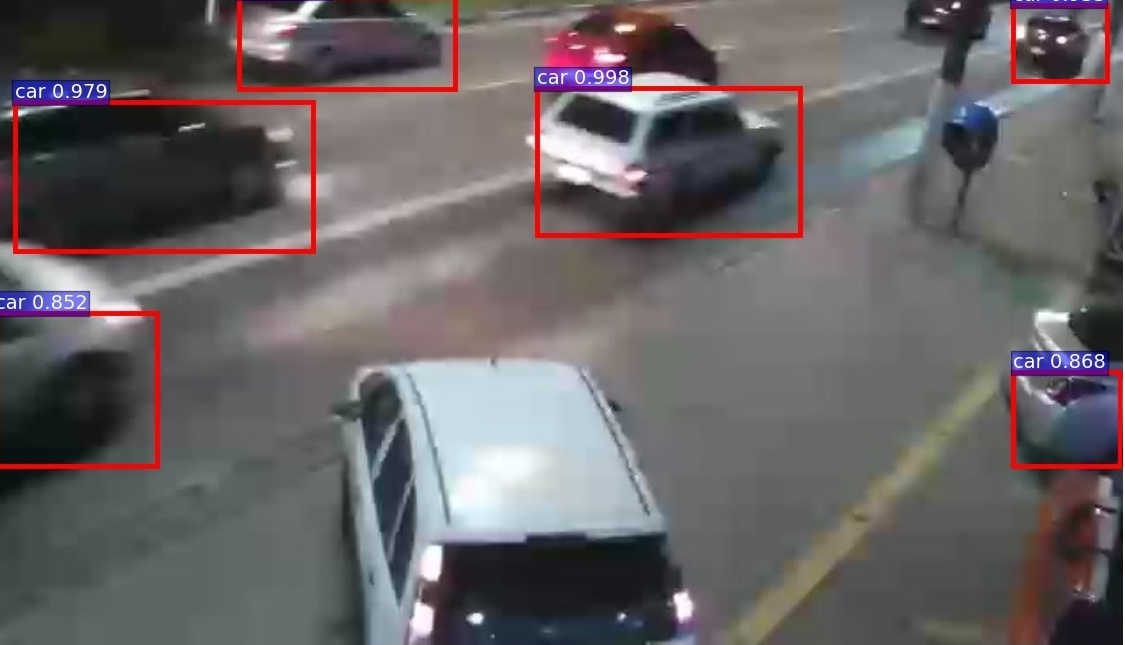} &
        \includegraphics[width=0.23\textwidth]{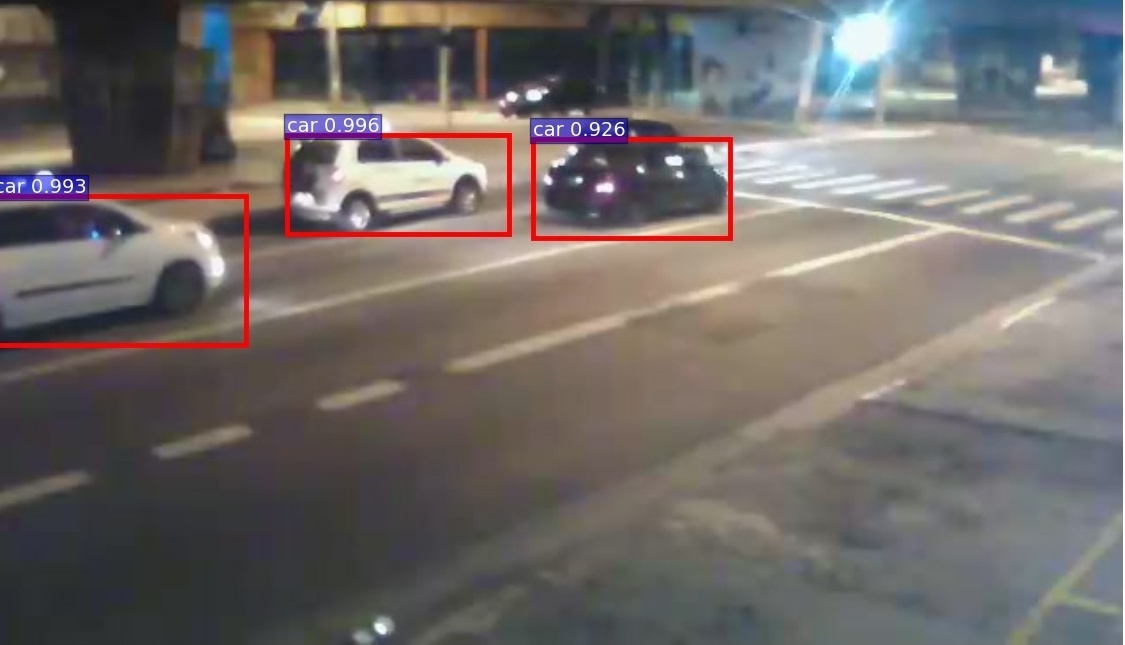} &
        \includegraphics[width=0.23\textwidth]{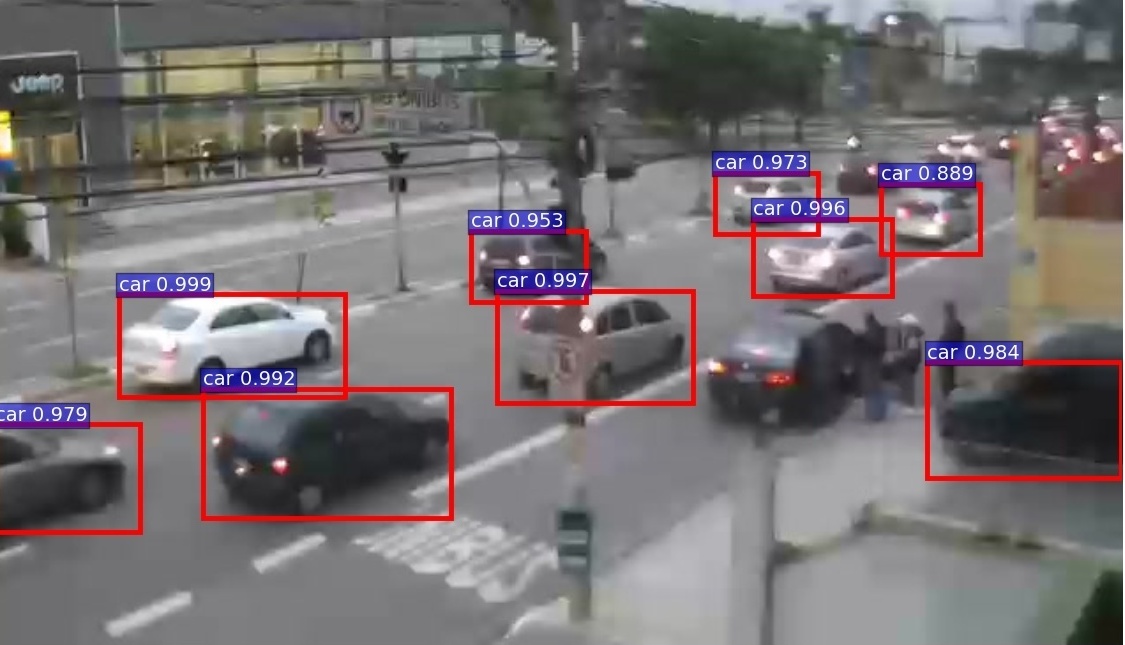} &
        \includegraphics[width=0.23\textwidth]{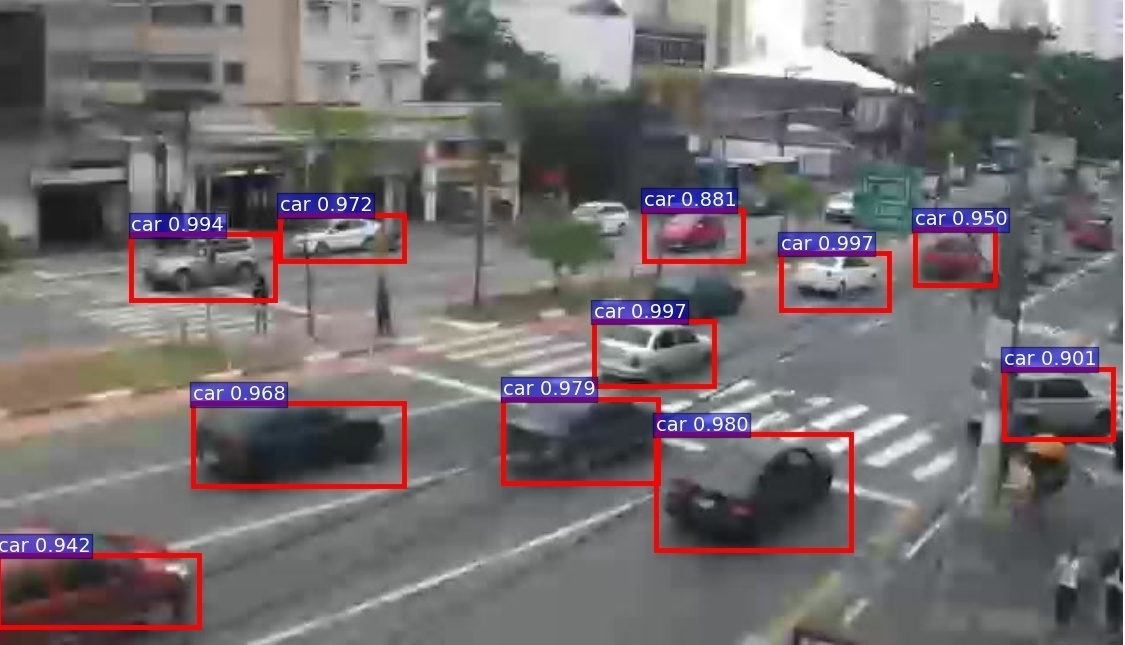} \\
        \includegraphics[width=0.23\textwidth]{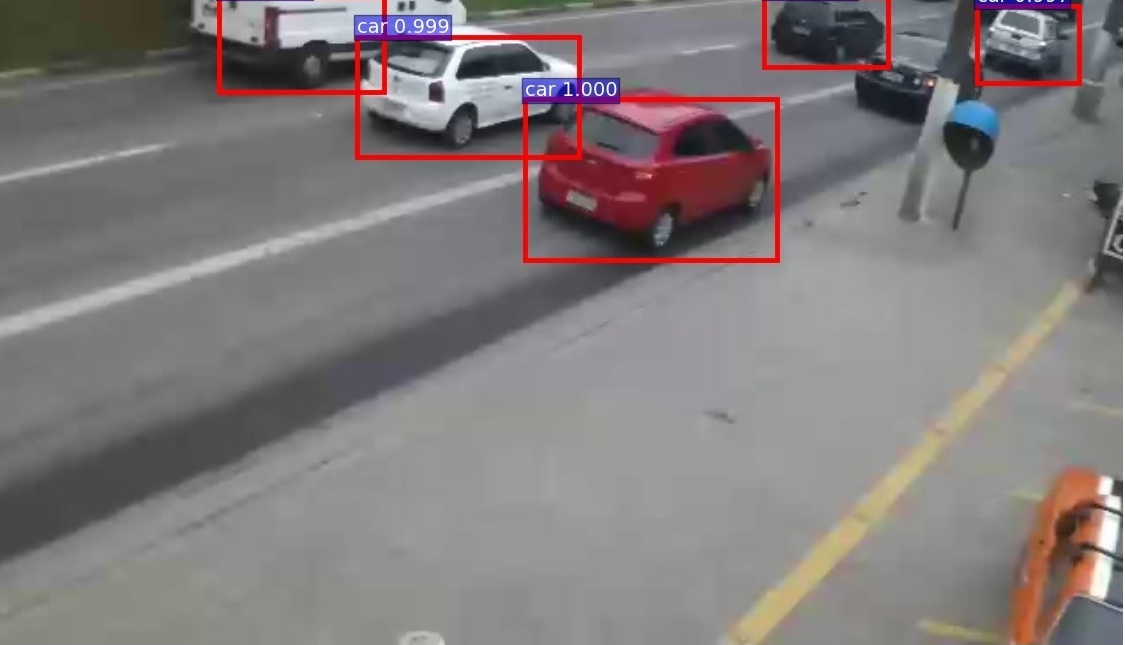} &
        \includegraphics[width=0.23\textwidth]{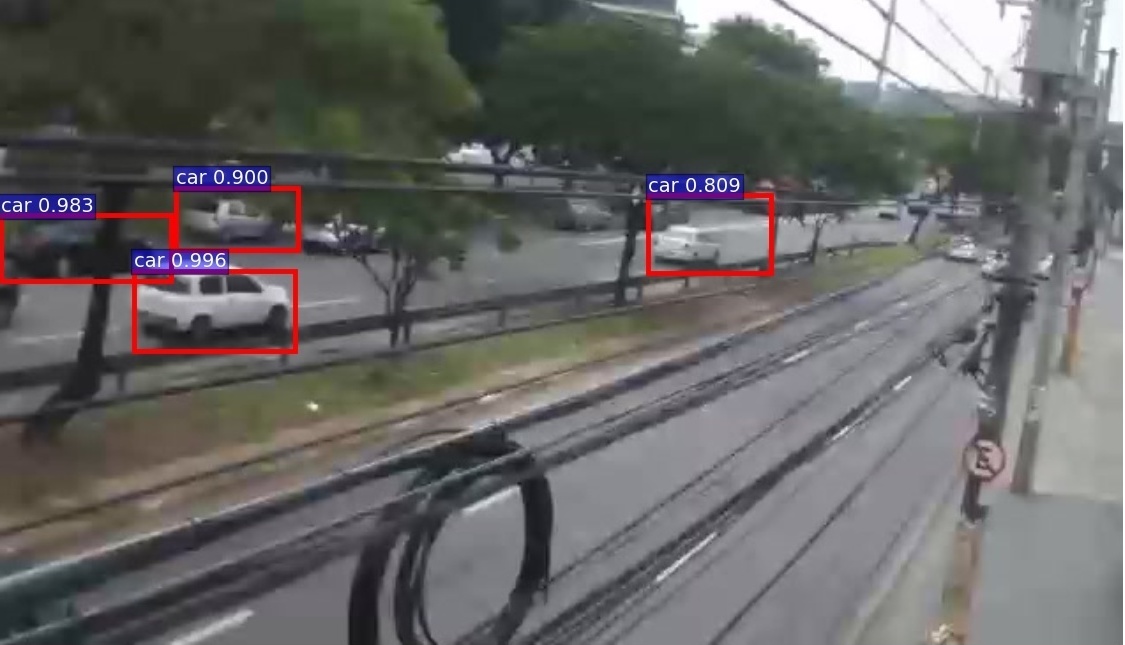} &
        \includegraphics[width=0.23\textwidth]{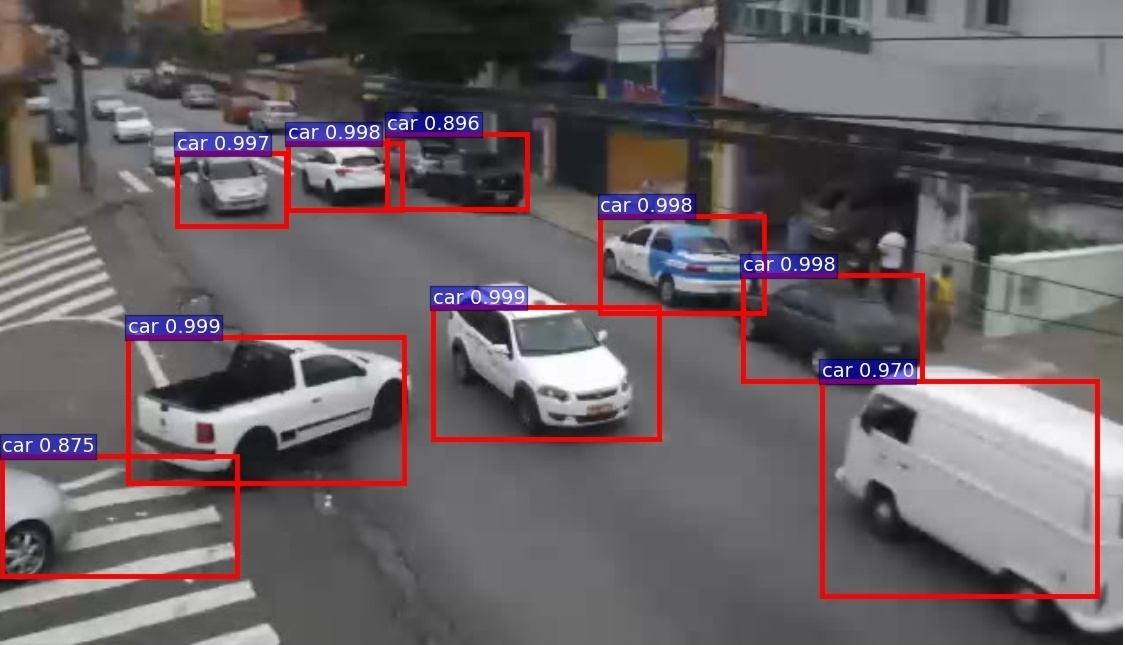} &
        \includegraphics[width=0.23\textwidth]{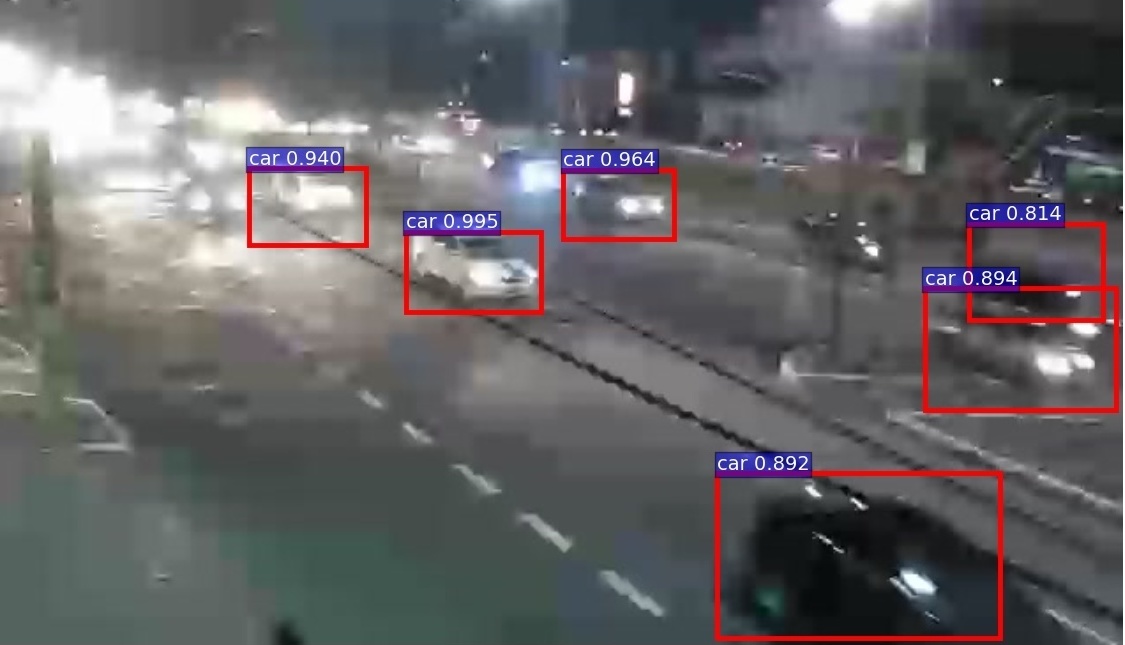} 
        \label{tab:mosaic}
        \end{tabular}
\end{table*}



\subsection{Validation 1: Uncontrolled weather}
In the first experiment, we created a detector of cars based on images from monitoring cameras. 24 cameras were continuously monitored during 6 days resulting in 768 hour of videos and 265.2 GB. The videos were systematically sampled at a 1/20 rate from the streaming to reduce the redundancy on the detections since time-varying information has not been explored in the current implementation (e.g. tracking, which can be easily incorporated in the proposed methodology). The 358,036 frames obtained were split into two groups. The trainning set of 300,000 frames ($\sim$84\%) and the test set of 58,036 ($\sim$16\%). We applied the DPM~\cite{felzenszwalb2010object} on the training set and a quality control step was performed on this stage. 557,036 cars were detected and used to fine-tune a RCNN VGG 16 layers~\cite{simonyan2014very} pre-trained with ImageNet, $RCNN_{imagenet}$. Thus a final detector, $RCNN_{all}$ was obtained. The second quality control stage was then performed. The quality control results are expressed in Table~\ref{tab:all}.

\begin{table*}[ht]
        \caption{Quality control over the Validation 1 and Validation 2.}
        \centering
        \begin{tabular}{clcccccc}
        \toprule
        &Detectors			&TP		& FP	&FN&Precision&Recall \\ \midrule
        \multirow{2}{*}{\begin{sideways}{\scriptsize Validation 1 }\end{sideways}}&WC ($DPM$)					&1366 	&127	&2205 		&91.5\%			&36.9\%	&\\
        &SC ($RCNN_{all}$)	&2638	&345 	&1060	&88.4\%		&71.3\%\\
        &$r.c._{WC}$ 		&		&		&	&-3.2\%	&+93.2\% \\
        \midrule
        \multirow{2}{*}{\begin{sideways}{\scriptsize Validation 2}\end{sideways}}
        &WC ($DPM$)				&914 	&115	&	2703&88.8\%		&25.2\%\\
        &SC ($RCNN_{rainy}$)	&1512	&449 	&	2105&77.1\%		&41.8\%\\
        &$r.c._{WC}$ 			&		&		&		&-12.5\% 	&+65.8\% \\
        \bottomrule
        \label{tab:all}
        \end{tabular}
\end{table*}

In the performance assessment of the $RCNN_{all}$, a relative change of precision of $rc(precision_{WC}) = -3.2\%$ and a relative change $rc(recall_{WC}) = 93.2\%$ on WC performance was obtained. The generated detector thus presents a significant increase in recall with the trade-off of losing a little precision.

\subsection{Validation 2: Rainy days}
In the second set of experiments, we created a detector of cars based on images of frames of rainy days from monitoring cameras.


We used 14 cameras during 2 days. Here as well, the raw videos were sampled at a 1/20 rate.  The 7,011 frames obtained were split into 6,000 (trainning set $\sim$85\%) and 1,011 (test set $\sim$15\%). We applied the DPM~\cite{felzenszwalb2010object} on the training set and a quality control step was performed on this stage. The 17,325 cars detected were used to fine-tune $RCNN_{all}$, the detector previously obtained with images from Camerite-all. A final detector $RCNN_{rainy}$ was then obtained. The second quality control stage was then performed.  Manual inspection was performed over a sample of 500 images from the test set, according to  the quality control step proposed in Section~\ref{sec:experimental}. The results of the quality control stages are expressed in Table~\ref{tab:all}. In the performance assessment of the $RCNN_{rainy}$, relative changes $rc(precision_{WC})=$~-12.5\% and $rc(recall_{WC})=$~65.8\% were obtained.


\begin{figure*}
    \centering
    \begin{subfigure}[b]{0.32\textwidth}
        \includegraphics[width=\textwidth]{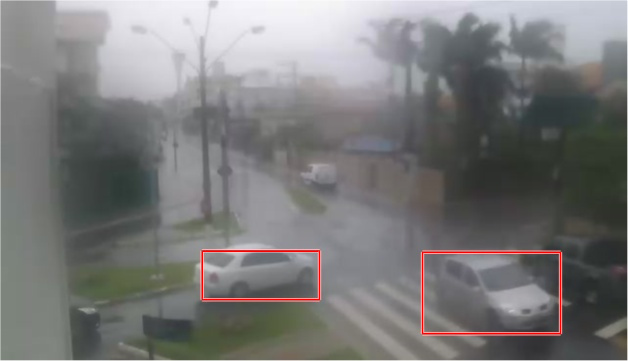}
    \end{subfigure}
    \begin{subfigure}[b]{0.32\textwidth}
        \includegraphics[width=\textwidth]{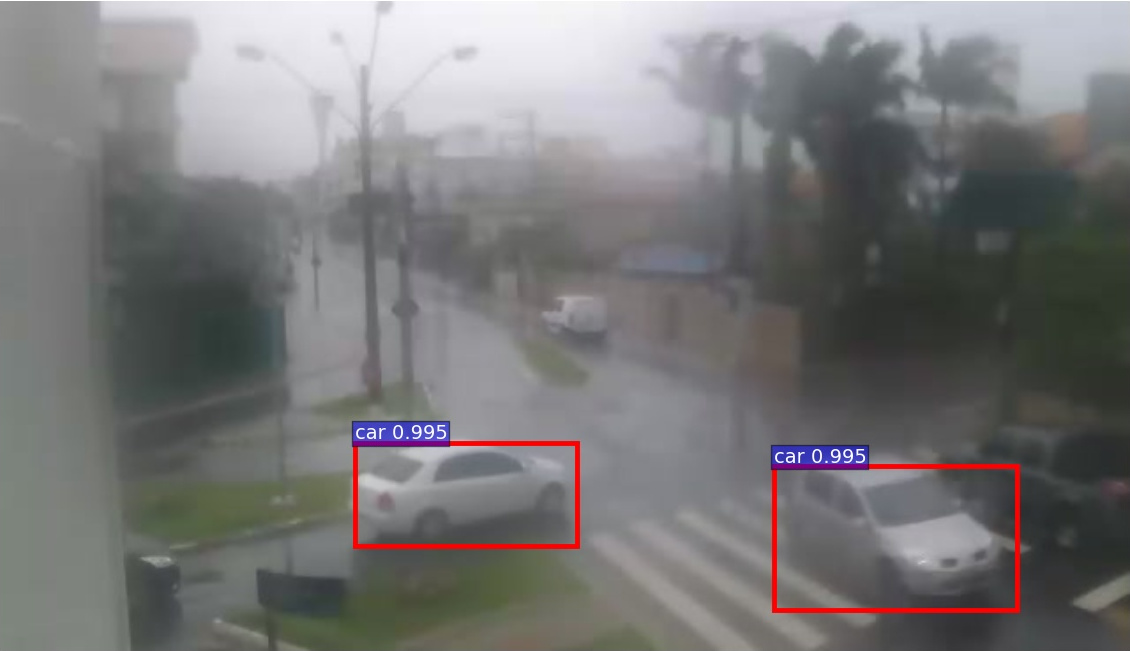}
    \end{subfigure}
    \begin{subfigure}[b]{0.32\textwidth}
        \includegraphics[width=\textwidth]{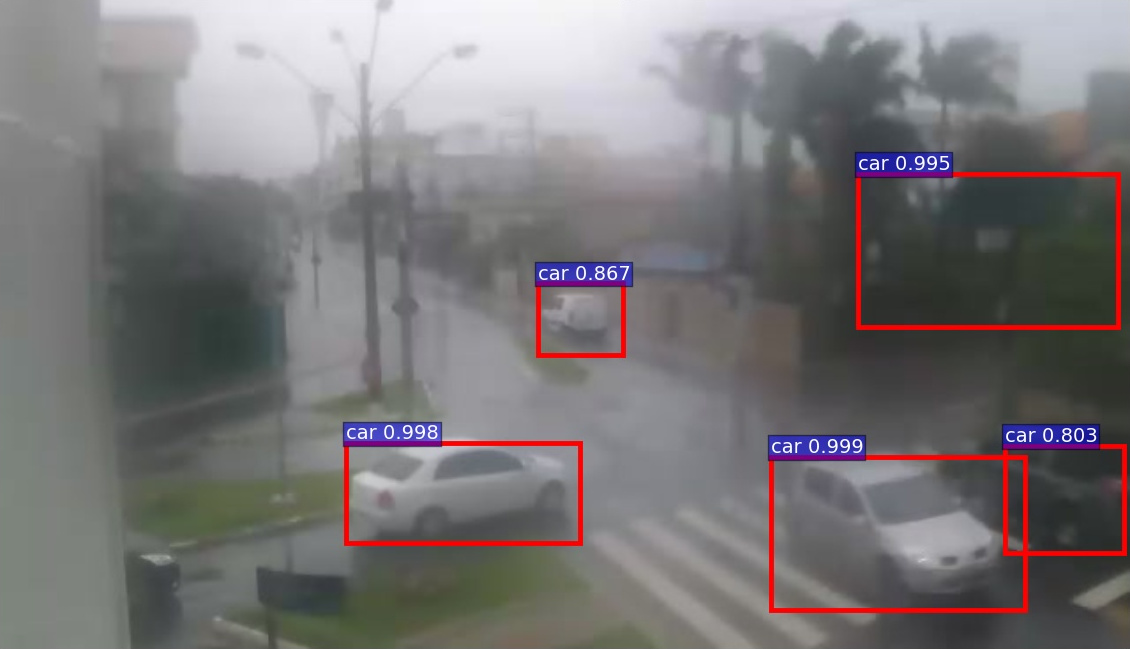}
    \end{subfigure}
    \caption{Comparison of the cars detections performed by the $DPM$, $RCNN_{all}$ and $RCNN_{rainy}$, respectively, from left to right.}
    \label{fig:detection}
\end{figure*}

\section{Discussion and concluding remarks}
\label{sec:conclusion}
 


This work presents the current state of a urban informatics framework that involves three data levels: Source data, City model and Knowledge. Our ongoing work focus on developing this framework with an implementation using real cases. The framework is based on:

\begin{itemize}
	\item Source data: city information should be collected as automatically as possible. We consider two types of data: visual and non-visual:
  \begin{itemize}
      \item Visual data: city maps, remote sensing, urban images/videos;
      \item Non-visual urban data: All types of socio-economic statistics available like education levels and information, violence, news information, traffic, etc.    
  \end{itemize}
	\item City model: representation of  different layers of city features. Different data structures have to be defined (e.g. images, networks,  textual and numerical records, etc). 
	\item Knowledge: The framework should help to address questions like \emph{How do cities evolve?}, \emph{How can they be compared?}. Such questions should be answered with the support of analytical methods suitable for each case provided by the framework.
\end{itemize}

Annotating large amounts of data is challenging. An option is to perform it manually, which is labor-intensive for big data. Alternative options include  hiring services like Amazon Mechanical Turk or Citizen Science~\cite{mturk2013}. However, new approaches to minimize human operation are desirable. In this scenario, the contribution  of this paper is the proposal of a methodology for generating object detectors with minimal manual annotation and quality control. The source data is initially processed by a a weak classifier and the resulting detections generated are used to fine-tune a new detector. In both steps, the user inspects a small sample of detections looking just on the true and false positives ratio. We validated it in the creation of a car detector for monitoring cameras that was able to produce a relative change on the precision and on the recall of the weak classifier of {-3.2} and {93.2}, respectively. Motivated by the urban problem, we performed the same pipeline using  rainy images and we got -12.5 and 65.8, respectively. These results show that the strong classifier presents a substantial improvement in recall with a small loss of precision.

For this paper we implemented one step of the collection and annotation of visual data in an urban environment. Next steps include the extension of our method to other urban elements such as pedestrians, buildings and roads. Then, other visual data sources besides city images will be explored including city maps and radar images. Following, non-visual data like from demography, traffic and violence information will be incorporated and, finally, we are going to create a city model to address the aforementioned urban issues.

\section*{Acknowledgments}

The authors would like to thank
FAPESP grants \#2015/22308-2, \#15/03475-5, \#	16/12077-6, \#	14/24918-0, CNPq, CAPES and NAP eScience - PRP - USP.

\bibliographystyle{plain}
\bibliography{main}

\begin{thebibliography}{10}

\bibitem{abu2012learning}
Yaser~S Abu-Mostafa, Malik Magdon-Ismail, and Hsuan-Tien Lin.
\newblock {\em Learning from data}, volume~4.
\newblock AMLBook New York, NY, USA:, 2012.

\bibitem{ali2007lagrangian}
Saad Ali and Mubarak Shah.
\newblock A lagrangian particle dynamics approach for crowd flow segmentation
  and stability analysis.
\newblock In {\em Computer Vision and Pattern Recognition, 2007. CVPR'07. IEEE
  Conference on}, pages 1--6. IEEE, 2007.

\bibitem{statisticsForComputer}
Michael Baron.
\newblock Probability and statistics for computer scientists.
\newblock pages 256--257. CRC Press, 2 edition, 2014.

\bibitem{brostow2009semantic}
Gabriel~J Brostow, Julien Fauqueur, and Roberto Cipolla.
\newblock Semantic object classes in video: A high-definition ground truth
  database.
\newblock {\em Pattern Recognition Letters}, 30(2):88--97, 2009.

\bibitem{camerite2016}
Camerite.
\newblock {\em (http://www.camerite.com)}.
\newblock {Camerite Publicidade e Monitoramente Ltda.}, Santa Catarina, Brazil,
  Last accessed March 2017.

\bibitem{carbonetto2008learning}
Peter Carbonetto, Gyuri Dork{\'o}, Cordelia Schmid, Hendrik K{\"u}ck, and Nando
  De~Freitas.
\newblock Learning to recognize objects with little supervision.
\newblock {\em International Journal of Computer Vision}, 77(1-3):219--237,
  2008.

\bibitem{carreira2010constrained}
Joao Carreira and Cristian Sminchisescu.
\newblock Constrained parametric min-cuts for automatic object segmentation.
\newblock In {\em Computer Vision and Pattern Recognition (CVPR), 2010 IEEE
  Conference on}, pages 3241--3248. IEEE, 2010.

\bibitem{cheng2014bing}
Ming-Ming Cheng, Ziming Zhang, Wen-Yan Lin, and Philip Torr.
\newblock Bing: Binarized normed gradients for objectness estimation at 300fps.
\newblock In {\em Proceedings of the IEEE Conference on Computer Vision and
  Pattern Recognition}, pages 3286--3293, 2014.

\bibitem{cortes1995support}
Corinna Cortes and Vladimir Vapnik.
\newblock Support-vector networks.
\newblock {\em Machine learning}, 20(3):273--297, 1995.

\bibitem{dalal2005histograms}
Navneet Dalal and Bill Triggs.
\newblock Histograms of oriented gradients for human detection.
\newblock In {\em Computer Vision and Pattern Recognition, 2005. CVPR 2005.
  IEEE Computer Society Conference on}, volume~1, pages 886--893. IEEE, 2005.

\bibitem{united2016air}
United States Environment Protection Agency~Air Data.
\newblock {\em (https://www3.epa.gov/airdata/ad\_data\_daily.html)}.
\newblock United States Environment Protection Agency, Last accessed March
  2017.

\bibitem{dollar2012pedestrian}
Piotr Dollar, Christian Wojek, Bernt Schiele, and Pietro Perona.
\newblock Pedestrian detection: An evaluation of the state of the art.
\newblock {\em Pattern Analysis and Machine Intelligence, IEEE Transactions
  on}, 34(4):743--761, 2012.

\bibitem{everingham2010pascal}
Mark Everingham, Luc Van~Gool, Christopher~KI Williams, John Winn, and Andrew
  Zisserman.
\newblock The pascal visual object classes (voc) challenge.
\newblock {\em International journal of computer vision}, 88(2):303--338, 2010.

\bibitem{felzenszwalb2010object}
Pedro~F Felzenszwalb, Ross~B Girshick, David McAllester, and Deva Ramanan.
\newblock Object detection with discriminatively trained part-based models.
\newblock {\em Pattern Analysis and Machine Intelligence, IEEE Transactions
  on}, 32(9):1627--1645, 2010.

\bibitem{geiger2013vision}
Andreas Geiger, Philip Lenz, Christoph Stiller, and Raquel Urtasun.
\newblock Vision meets robotics: The kitti dataset.
\newblock {\em International Journal of Robotics Research (IJRR)}, 2013.

\bibitem{girshick2014rich}
Ross Girshick, Jeff Donahue, Trevor Darrell, and Jitendra Malik.
\newblock Rich feature hierarchies for accurate object detection and semantic
  segmentation.
\newblock In {\em Proceedings of the IEEE conference on computer vision and
  pattern recognition}, pages 580--587, 2014.

\bibitem{griffin2007caltech}
Gregory Griffin, Alex Holub, and Pietro Perona.
\newblock Caltech-256 object category dataset.
\newblock 2007.

\bibitem{hariharan2012discriminative}
Bharath Hariharan, Jitendra Malik, and Deva Ramanan.
\newblock Discriminative decorrelation for clustering and classification.
\newblock In {\em Computer Vision--ECCV 2012}, pages 459--472. Springer, 2012.

\bibitem{hosang2015makes}
Jan Hosang, Rodrigo Benenson, Piotr Doll{\'a}r, and Bernt Schiele.
\newblock What makes for effective detection proposals?
\newblock {\em Pattern Analysis and Machine Intelligence, IEEE Transactions
  on}, 2015.

\bibitem{hosmer2007ilids}
Paul Hosmer.
\newblock i-lids dataset for avss 2007.
\newblock In {\em Advanced Video and Signal based Surveillance}. IEEE, 2007.

\bibitem{earthcam}
Earthcam Inc.
\newblock {\em (https://www.earthcam.com)}.
\newblock EarthCam, Last accessed March 2017.

\bibitem{googlemaps}
Google Inc.
\newblock {\em (https://maps.google.com)}.
\newblock Google Maps, Last accessed March 2017.

\bibitem{kalal2012tracking}
Zdenek Kalal, Krystian Mikolajczyk, and Jiri Matas.
\newblock Tracking-learning-detection.
\newblock {\em IEEE transactions on pattern analysis and machine intelligence},
  34(7):1409--1422, 2012.

\bibitem{krause20133d}
Jonathan Krause, Michael Stark, Jia Deng, and Li~Fei-Fei.
\newblock 3d object representations for fine-grained categorization.
\newblock In {\em Proceedings of the IEEE International Conference on Computer
  Vision Workshops}, pages 554--561, 2013.

\bibitem{krizhevsky2012imagenet}
Alex Krizhevsky, Ilya Sutskever, and Geoffrey~E Hinton.
\newblock Imagenet classification with deep convolutional neural networks.
\newblock In {\em Advances in neural information processing systems}, pages
  1097--1105, 2012.

\bibitem{le1991mpeg}
Didier Le~Gall.
\newblock Mpeg: A video compression standard for multimedia applications.
\newblock {\em Communications of the ACM}, 34(4):46--58, 1991.

\bibitem{lichman2013uci}
Moshe Lichman.
\newblock {UCI} machine learning repository, 2013.

\bibitem{lin2014microsoft}
Tsung-Yi Lin, Michael Maire, Serge Belongie, James Hays, Pietro Perona, Deva
  Ramanan, Piotr Doll{\'a}r, and C~Lawrence Zitnick.
\newblock Microsoft coco: Common objects in context.
\newblock In {\em Computer Vision--ECCV 2014}, pages 740--755. Springer, 2014.

\bibitem{lowe1999object}
David~G Lowe.
\newblock Object recognition from local scale-invariant features.
\newblock In {\em Computer vision, 1999. The proceedings of the seventh IEEE
  international conference on}, volume~2, pages 1150--1157. Ieee, 1999.

\bibitem{misra2015watch}
Ishan Misra, Abhinav Shrivastava, and Martial Hebert.
\newblock Watch and learn: Semi-supervised learning for object detectors from
  video.
\newblock In {\em Proceedings of the IEEE Conference on Computer Vision and
  Pattern Recognition}, pages 3593--3602, 2015.

\bibitem{oh2011large}
Sangmin Oh, Anthony Hoogs, Amitha Perera, Naresh Cuntoor, Chia-Chih Chen,
  Jong~Taek Lee, Saurajit Mukherjee, JK~Aggarwal, Hyungtae Lee, Larry Davis,
  et~al.
\newblock A large-scale benchmark dataset for event recognition in surveillance
  video.
\newblock In {\em Computer Vision and Pattern Recognition (CVPR), 2011 IEEE
  Conference on}, pages 3153--3160. IEEE, 2011.

\bibitem{ozuysal2009pose}
Mustafa Ozuysal, Vincent Lepetit, and Pascal Fua.
\newblock Pose estimation for category specific multiview object localization.
\newblock In {\em Computer Vision and Pattern Recognition, 2009. CVPR 2009.
  IEEE Conference on}, pages 778--785. IEEE, 2009.

\bibitem{insecam}
Insecam Project.
\newblock {\em (https://www.insecam.org)}.
\newblock Insecam, Last accessed March 2017.

\bibitem{ren2015faster}
Shaoqing Ren, Kaiming He, Ross Girshick, and Jian Sun.
\newblock Faster r-cnn: Towards real-time object detection with region proposal
  networks.
\newblock In {\em Advances in Neural Information Processing Systems}, pages
  91--99, 2015.

\bibitem{russakovsky2015imagenet}
Olga Russakovsky, Jia Deng, Hao Su, Jonathan Krause, Sanjeev Satheesh, Sean Ma,
  Zhiheng Huang, Andrej Karpathy, Aditya Khosla, Michael Bernstein, et~al.
\newblock Imagenet large scale visual recognition challenge.
\newblock {\em International Journal of Computer Vision}, 115(3):211--252,
  2015.

\bibitem{schmidhuber2015deep}
J{\"u}rgen Schmidhuber.
\newblock Deep learning in neural networks: An overview.
\newblock {\em Neural Networks}, 61:85--117, 2015.

\bibitem{simonyan2014very}
Karen Simonyan and Andrew Zisserman.
\newblock Very deep convolutional networks for large-scale image recognition.
\newblock {\em arXiv preprint arXiv:1409.1556}, 2014.

\bibitem{soomro2012ucf101}
Khurram Soomro, Amir~Roshan Zamir, and Mubarak Shah.
\newblock Ucf101: A dataset of 101 human actions classes from videos in the
  wild.
\newblock {\em arXiv preprint arXiv:1212.0402}, 2012.

\bibitem{sorokin2008utility}
Alexander Sorokin and David Forsyth.
\newblock Utility data annotation with amazon mechanical turk.
\newblock 2008.

\bibitem{stockhammer2011dynamic}
Thomas Stockhammer.
\newblock Dynamic adaptive streaming over http--: standards and design
  principles.
\newblock In {\em Proceedings of the second annual ACM conference on Multimedia
  systems}, pages 133--144. ACM, 2011.

\bibitem{szeliski2010computer}
Richard Szeliski.
\newblock {\em Computer vision: algorithms and applications}.
\newblock Springer Science \& Business Media, 2010.

\bibitem{mturk2013}
Mechanical Turk.
\newblock {\em (http://www.mturk.com)}.
\newblock Amazon Web Services, Amazon Inc., 2013.

\bibitem{van2011segmentation}
Koen~EA Van~de Sande, Jasper~RR Uijlings, Theo Gevers, and Arnold~WM Smeulders.
\newblock Segmentation as selective search for object recognition.
\newblock In {\em Computer Vision (ICCV), 2011 IEEE International Conference
  on}, pages 1879--1886. IEEE, 2011.

\bibitem{postgresql1996}
{PostgreSQL} version 9.4.6.
\newblock {\em (http://www.postgresql.org)}.
\newblock PostgreSQL Global Development Group, California, United States, 2010.

\bibitem{vezzani2008visor}
Roberto Vezzani and Rita Cucchiara.
\newblock Visor: Video surveillance on-line repository for annotation
  retrieval.
\newblock In {\em Multimedia and Expo, 2008 IEEE International Conference on},
  pages 1281--1284. IEEE, 2008.

\bibitem{viola2001rapid}
Paul Viola and Michael Jones.
\newblock Rapid object detection using a boosted cascade of simple features.
\newblock In {\em Computer Vision and Pattern Recognition, 2001. CVPR 2001.
  Proceedings of the 2001 IEEE Computer Society Conference on}, volume~1, pages
  I--511. IEEE, 2001.

\bibitem{wang2007unsupervised}
Xiaogang Wang, Xiaoxu Ma, and Eric Grimson.
\newblock Unsupervised activity perception by hierarchical bayesian models.
\newblock In {\em Computer Vision and Pattern Recognition, 2007. CVPR'07. IEEE
  Conference on}, pages 1--8. IEEE, 2007.

\bibitem{yang2015large}
Linjie Yang, Ping Luo, Chen Change~Loy, and Xiaoou Tang.
\newblock A large-scale car dataset for fine-grained categorization and
  verification.
\newblock In {\em Proceedings of the IEEE Conference on Computer Vision and
  Pattern Recognition}, pages 3973--3981, 2015.

\bibitem{zhang2006rain}
Xiaopeng Zhang, Hao Li, Yingyi Qi, Wee~Kheng Leow, and Teck~Khim Ng.
\newblock Rain removal in video by combining temporal and chromatic properties.
\newblock In {\em Multimedia and Expo, 2006 IEEE International Conference on},
  pages 461--464. IEEE, 2006.

\bibitem{zitnick2014edge}
C~Lawrence Zitnick and Piotr Doll{\'a}r.
\newblock Edge boxes: Locating object proposals from edges.
\newblock In {\em Computer Vision--ECCV 2014}, pages 391--405. Springer, 2014.

\end{thebibliography}
\end{document}